\documentclass[10pt,twocolumn,letterpaper]{article}

\usepackage[pagenumbers]{iccv} 

\definecolor{iccvblue}{rgb}{0.21,0.49,0.74}
\definecolor{calpolypomonagreen}{rgb}{0.12, 0.3, 0.17}
\usepackage[pagebackref,breaklinks,colorlinks,allcolors=iccvblue]{hyperref}
\usepackage[font=small, skip=4pt]{caption}
\usepackage{amsmath}
\usepackage{upgreek}
\usepackage{adjustbox}
\usepackage{xcolor}
\usepackage{color, colortbl}
\usepackage{booktabs}
\usepackage{arydshln}
\usepackage{multirow}
\usepackage{algorithm}
\usepackage{algpseudocode}
\usepackage{graphicx}
\usepackage{tabularx}
\usepackage{animate}

\hypersetup{
    urlcolor=calpolypomonagreen,
}

\def\paperID{6552} 
\def\confName{ICCV}
\def\confYear{2025}

\newcommand\norm[1]{\left\lVert#1\right\rVert}

\newcommand{\xb}{{\boldsymbol x}}
\newcommand{\mb}{{\boldsymbol m}}

\newcommand{\zb}{{\boldsymbol z}}

\newcommand{\vb}{{\boldsymbol v}}

\newcommand{\cb}{{\boldsymbol c}}

\newcommand{\epsilonb}{{\boldsymbol \epsilon}}

\newcommand{\Eb}{{\mathbb E}}

\usepackage{amsthm}

\newcommand\Mark[1]{\textsuperscript#1}

\title{
 Reangle-A-Video: 4D Video Generation as Video-to-Video Translation
}

\author{
    Hyeonho Jeong\Mark{1}\Mark{,}\Mark{2}\Mark{,}\Mark{*} \qquad\qquad 
    Suhyeon Lee\Mark{1}\Mark{,}\Mark{*} \qquad\qquad 
    Jong Chul Ye\Mark{1}\\[0.04in]
    \Mark{1}KAIST \qquad
    \Mark{2}Adobe Research \qquad\\
    \texttt{\{hyeonho.jeong, suhyeon.lee, jong.ye\}@kaist.ac.kr} 
}

\begin{document}

\twocolumn[{
\renewcommand\twocolumn[1][]{#1}
\maketitle
\begin{center}
    \centering
    \captionsetup{type=figure}
    \includegraphics[width=\textwidth]{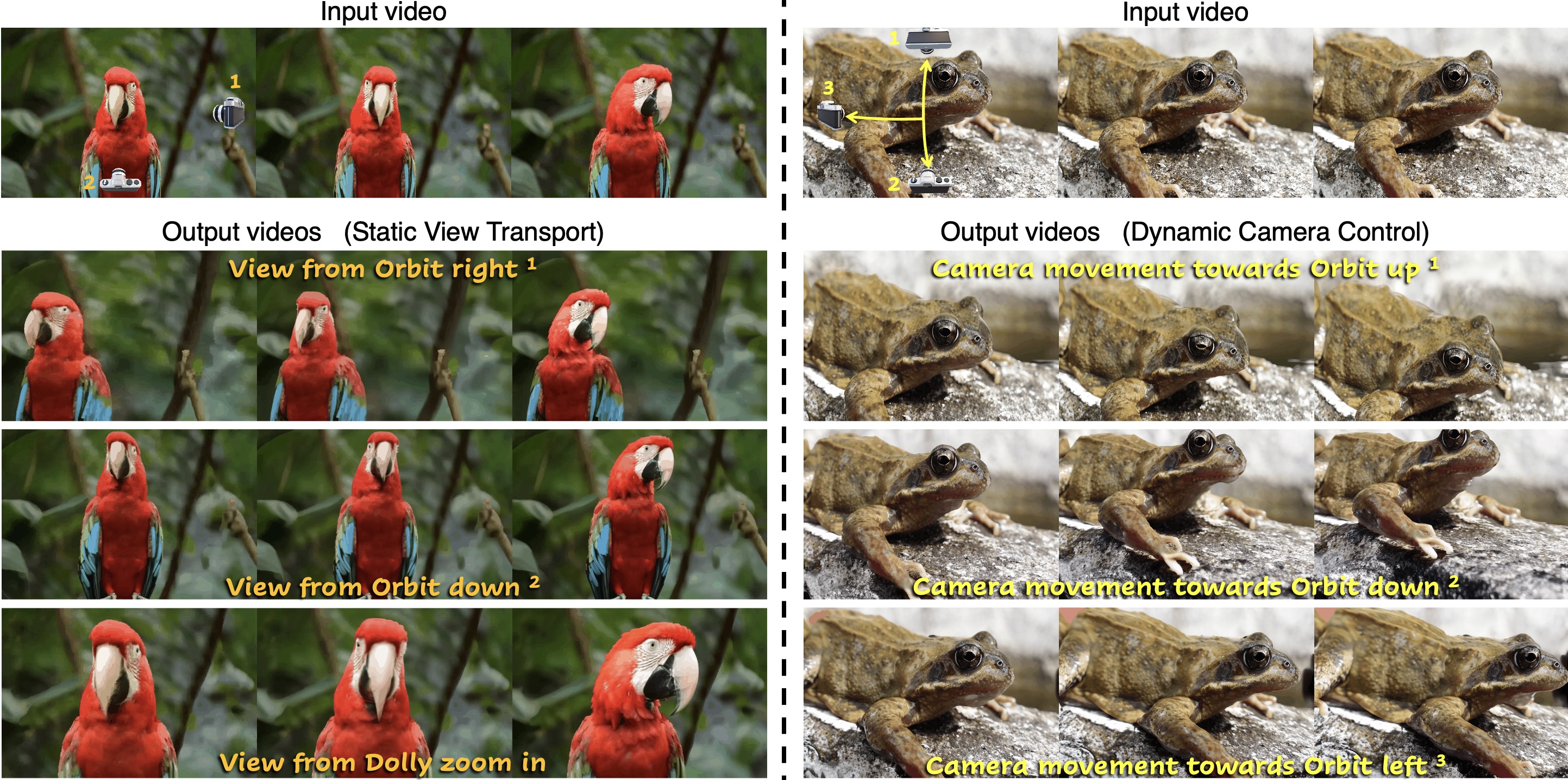}
    \captionof{figure}
    {
        From a single monocular video of any scene, \texttt{Reangle-A-Video} generates synchronized videos from diverse camera viewpoints or movements without relying on any multi-view generative prior\textit{—using only single fine-tuning of a video generator}.
        The first row shows the input video, while the rows below present videos generated by Reangle-A-Video.
        (Left): Static view transport results. 
        (Right): Dynamic camera control results.
        Full video examples are available on our project page: \textit{
        \href{https://hyeonho99.github.io/reangle-a-video/}{hyeonho99.github.io/reangle-a-video}
        }
    }
\label{fig: teaser}
\end{center}
}]

\begin{abstract}
We introduce Reangle-A-Video,  a unified framework for generating synchronized multi-view videos from a single input video.
Unlike mainstream approaches that train multi-view video diffusion models on large-scale 4D datasets, our method reframes the multi-view video generation task as video-to-videos translation, leveraging publicly available image and video diffusion priors.
In essence, Reangle-A-Video operates in two stages. 
(1) Multi-View Motion Learning:
An image-to-video diffusion transformer is synchronously fine-tuned in a self-supervised manner to distill view-invariant motion from a set of warped videos.
(2) Multi-View Consistent Image-to-Images Translation:
The first frame of the input video is warped and inpainted into various camera perspectives under an inference-time cross-view consistency guidance using DUSt3R, generating multi-view consistent starting images.
Extensive experiments on static view transport and dynamic camera control show that Reangle-A-Video surpasses existing methods, establishing a new solution for multi-view video generation. 
We will publicly release our code and data.
\let\thefootnote\relax\footnotetext{$^*$ indicates equal contribution}
\end{abstract}

\setlength{\belowcaptionskip}{-4pt}
\begin{table*}[t]
\centering
    \newcommand{\numColumns}{6}
    \newcommand{\columnSpacing}{0.03cm}
    \newcommand{\fps}{12}
    \newcommand{\newcolumnwidth}{\textwidth}
    \newcommand{\colwidth}{\dimexpr (\newcolumnwidth - 6*\columnSpacing - \arrayrulewidth)/6 \relax}
    \newcommand{\groupTwoWidth}{\dimexpr 2\colwidth+\columnSpacing \relax}

    \begin{tabular}{
        @{}*{\numColumns}{p{\dimexpr(\newcolumnwidth-\columnSpacing*(\numColumns-1))/\numColumns}@{\hspace{\columnSpacing}}}@{}
    }
        \multicolumn{1}{c}{\footnotesize Input video} & 
        \multicolumn{2}{c}{\footnotesize Static view transport results} &
        \multicolumn{1}{c}{\footnotesize Input video} &
        \multicolumn{2}{c}{\footnotesize Dynamic camera control results}
         \\
    
        \animategraphics[loop, width=\linewidth]{\fps}{gallery/lion/input_jpeg/00}{00}{48} &
        \animategraphics[loop, width=\linewidth]{\fps}{gallery/lion/horizontal+8_jpeg/00}{00}{48} &
        \animategraphics[loop, width=\linewidth]{\fps}{gallery/lion/vertical+4_jpeg/00}{00}{48} &
        
        \animategraphics[loop, width=\linewidth]{\fps}{gallery/train/input_jpeg/00}{00}{48} &
        \animategraphics[loop, width=\linewidth]{\fps}{gallery/train/horizontal+0.2_jpeg/00}{00}{48} &
        \animategraphics[loop, width=\linewidth]{\fps}{gallery/train/zoomout+0.2_jpeg/00}{00}{48} \\
    \end{tabular}

    \vspace{-0.15cm}

    \begin{tabular}{
        @{}*{\numColumns}{p{\dimexpr(\newcolumnwidth-\columnSpacing*(\numColumns-1))/\numColumns}@{\hspace{\columnSpacing}}}@{}
    }
        \animategraphics[loop, width=\linewidth]{\fps}{gallery/blackswan/input_jpeg/00}{00}{48} &
        \animategraphics[loop, width=\linewidth]{\fps}{gallery/blackswan/horizontal+10_jpeg/00}{00}{48} &
        \animategraphics[loop, width=\linewidth]{\fps}{gallery/blackswan/horizontal-10_jpeg/00}{00}{48} &
        
        \animategraphics[loop, width=\linewidth]{\fps}{gallery/lama/input_jpeg/00}{00}{48} &
        \animategraphics[loop, width=\linewidth]{\fps}{gallery/lama/zoomin-0.4_jpeg/00}{00}{48} &
        \animategraphics[loop, width=\linewidth]{\fps}{gallery/lama/vertical-0.25_jpeg/00}{00}{48}
    \end{tabular}
    \vspace*{-2mm}
    \captionof{figure}{
        \textbf{Qualitative results on \textit{static view transport (left)} \& \textit{dynamic camera control (right)}}. Click with Acrobat Reader to play videos.
    }
    \vspace*{-2mm}
    \label{fig: gallery}
\end{table*}

\section{Introduction}
\label{sec:intro}
Diffusion-based video generators \cite{zhang2024show, wang2024lavie, blattmann2023stable, jeong2024track4gen, bar2024lumiere, videoworldsimulators2024, yang2024cogvideox, polyak2024moviegencastmedia, kong2024hunyuanvideo} are rapidly advancing, enabling the generation of visually rich and dynamic videos from text or visual inputs.
Recent progress in video generative models highlights the growing need for user control over object appearance \cite{jiang2024videobooth, wei2024dreamvideo2, kwon2024tweediemix, chefer2024still}, object motion \cite{jeong2024vmc, zhao2024motiondirector, yatim2024space, park2024spectral, jeong2024dreammotion, geng2024motion, burgert2025go}, and camera pose \cite{wang2024motionctrl, yang2024direct, he2024cameractrl, xu2024camco, you2024nvs, xiao2024trajectory, popov2025camctrl3d, zhang2024recapture}.
However, a video itself inherently captures only a partial perspective of the world, which exists as a dynamic 4D environment.

To obtain dynamic 4D generative priors, previous works \cite{li2024vivid, xie2024sv4d, liang2024diffusion4d, zuo2024videomv} have extended 3D or video generators into 4D generative models using rendered synthetic assets \cite{deitke2023objaverse, deitke2024objaverse} that focus on animated objects or rigged characters. 
As a result, these methods are limited to object-level 4D synthesis and fail to generalize to real-world scenes.
Recent studies have addressed this limitation by training foundational multi-view video diffusion models \cite{wu2024cat4d, bai2024syncammaster, watson2024controlling, wang20244real, van2024generative, shao2024360, sun2024dimensionx} on hybrid datasets combining indoor/outdoor static scenes \cite{zhou2018stereo, ling2024dl3dv, dai2017scannet}, general single-view videos, and realistic scenes rendered by simulation engines \cite{greff2022kubric, sanders2016introduction, deitke2023objaverse, deitke2024objaverse}.
Although promising, these approaches often face challenges in real-world scenarios \cite{van2024generative}, are limited to specific domains (e.g., human-centric synthesis \cite{shao2024360}), and most are not publicly available.
More importantly, most existing methods generate multi-view videos from image or text inputs, rather than from user-input videos.

To address this, here we present Reangle-A-Video, an alternative solution for synchronized multi-view video generation that does not require specialized multi-view generative priors.
Given any input video, we frame the task as \textit{video-to-videos translation}, capitalizing on publicly available image and video diffusion models.
Our unified framework supports both \textit{static view transport}—resimulating a video from target viewpoints—and \textit{dynamic camera control}, where the video gradually transitions to the target viewpoints. 
Both approaches offer six degrees of freedom (see Fig. \ref{fig: teaser} and \ref{fig: gallery}).

In essence, our approach is based on the decomposition of a dynamic 4D scene into view-specific appearance (\textit{the starting image}) and view-invariant motion (\textit{image-to-video generation}).
To capture view-invariant motion of the scene, we augment the training dataset with warped videos generated via repeated point-based warping of a single monocular video—these warped videos provide strong hints about the camera's perspectives.
We then fine-tune a pre-trained image-to-video model \cite{yang2024cogvideox} using a synchronized few-shot training strategy building on a masked diffusion loss \cite{avrahami2023break, zhang2024recapture, chou2024generating}.
After training, dynamic camera control over video (Fig.\ref{fig: teaser},\ref{fig: gallery}-\textit{right}) is achieved by generating videos using the original first frame input with text that specifies the desired camera movement.
On the other hand, static view transport of a video requires viewpoint-transported starting images.
To achieve this, our method addresses several key technical challenges.
First, we generate the starting images by inpainting warped images with image diffusion prior \cite{flux}. Second,
inspired by test-time compute scaling \cite{wallace2023end, yeh2024training, kim2024free}, we enforce multi-view consistency in \textit{inference-time}, using a stochastic control guidance with an off-the-shelf multi-view stereo reconstruction network \cite{wang2024dust3r}.
We demonstrate the effectiveness of our approach on a variety of real-world scenes and metrics.
\vspace{-1mm}

\vspace{-2pt}
\section{Related Work}
\label{sec:formatting}
\vspace{-2pt}
\noindent\textbf{Denoising-based Image and Video Generation.}
The introduction of Diffusion Transformers (DiT) \cite{peebles2023scalable} revolutionized image and video generation by replacing the traditional U-Net with a transformer-based backbone. 
Combined with larger-scale, high-quality, curated datasets, this shift enabled more efficient and scalable diffusion generators.
In image generation, the PixArt family \cite{chen2023pixart, chen2024pixart, chen2024pixartdelta} extended DiT to text-to-image synthesis, showcasing its versatility. Stable Diffusion 3 \cite{esser2024scaling} and Flux \cite{flux} further advanced the field with Multi-modal Diffusion Transformers (MM-DiT), enabling bidirectional text-image interactions.
Following this trend, recent video diffusion models also integrate MM-DiT with spatio-temporal VAEs, achieving high-quality, long-form video generation.
In our work, we adopt Flux \cite{flux} for image diffusion and CogVideoX \cite{yang2024cogvideox} for video diffusion, both built on the MM-DiT architecture.

\begin{figure*}[!t]
    \centering
    \includegraphics[width=\textwidth]
    {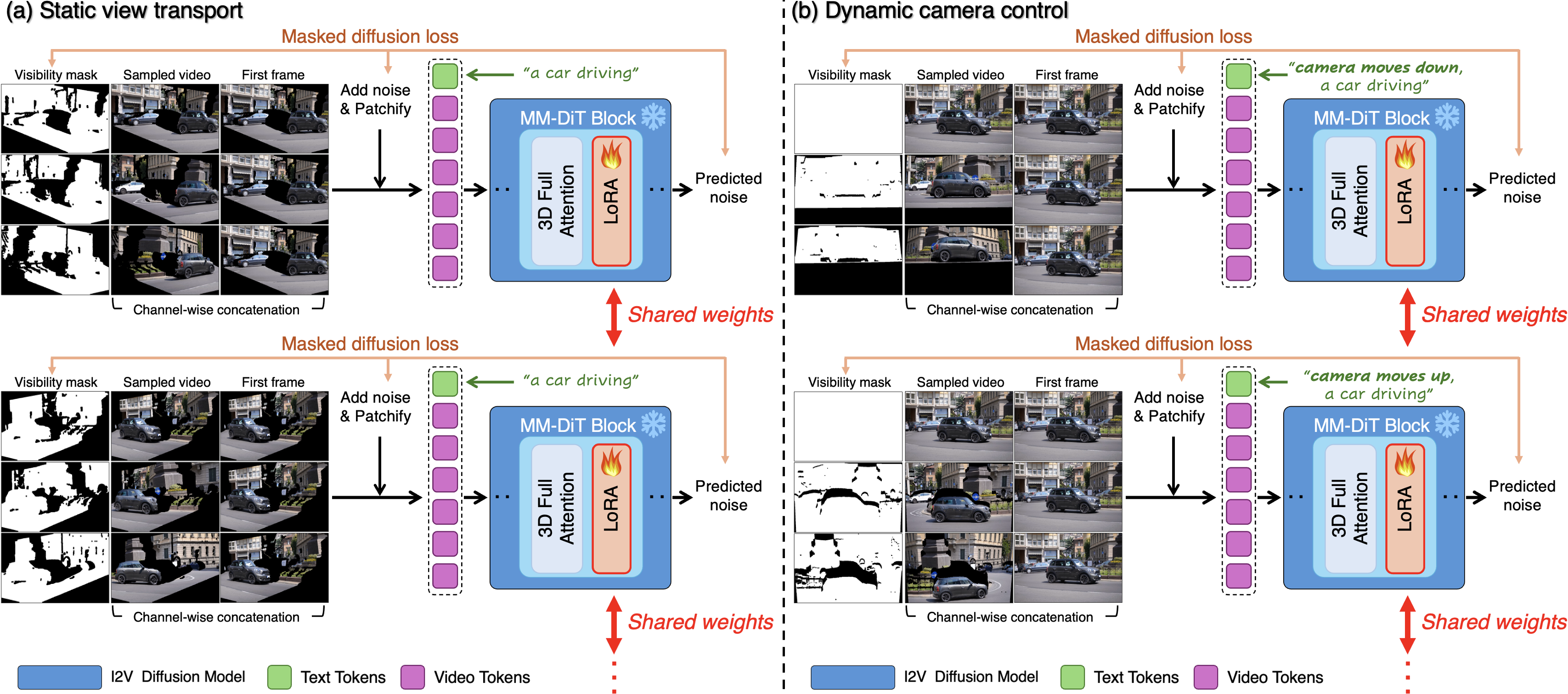}
    \caption{
    Multi-view motion learning pipelines for \textbf{(a) Static view transport} and \textbf{(b) Dynamic camera control.}
    For both tasks, we distill view-robust motion of the underlying scene to a pre-trained MM-DiT video model \cite{yang2024cogvideox}, using all visible pixels within the sampled videos.
    This few-shot, self-supervised training optimizes only the LoRA layers \cite{hu2022lora, ryu2023low}, enabling lightweight training.
    }
    \vspace{-2mm}
    \label{fig: method-motion-learning}
\end{figure*}

\noindent\textbf{Multi-View Video Generation.}
Previous efforts on multi-view (4D) video generation \cite{li2024vivid, xie2024sv4d, liang2024diffusion4d, zuo2024videomv} have mainly focused on synthesizing 4D assets—animated objects and rigged characters \cite{deitke2023objaverse, deitke2024objaverse}.
A common approach is to combine motion priors from video diffusion models with 3D priors from multi-view image models and train on curated, synthetic datasets of dynamic 3D objects \cite{li2024vivid, xie2024sv4d}.
While promising, these methods are limited to object-level 4D synthesis—rendered animations of single objects or animals—and produce multi-view videos with a static camera pose per video, failing to generalize to real-world videos and arbitrary viewpoints.
More recently, to capture open-domain real-world scenes, several works \cite{kuang2024collaborative, bai2024syncammaster, wu2024cat4d, wang20244real, sun2024dimensionx, van2024generative} have built foundational multi-view video generative priors by training on multi-view images/videos of real-world static scenes \cite{zhou2018stereo, ling2024dl3dv}, general single-view video datasets, and photorealistic scenes rendered by various engines \cite{deitke2023objaverse, greff2022kubric, sanders2016introduction}. 
These efforts aim to overcome the scarcity of dynamic, in-the-wild multi-view video data.

In contrast, our work reinterprets multi-view video generation as a \textit{video-to-multi-view-videos} translation task.
We demonstrate that synchronized few-shot fine-tuning of a video diffusion prior with visible pixels across arbitrary views is sufficient for multi-view video generation in real-world settings, without the need for expensive 4D diffusion models trained on large datasets.

\noindent\textbf{Camera Controllable Video Generation.}
Our framework is also related to video generation methods that incorporate controllable camera trajectories \cite{wang2024motionctrl, he2024cameractrl, popov2025camctrl3d, yang2024direct, bahmani2024ac3d, xu2024camco, hou2024training, kuang2025collaborative, li2025realcam, xu2024cavia, van2024generative, yu2024viewcrafter, zhang2024recapture, cheong2024boosting, you2024nvs, xiao2024trajectory}.
Most approaches fine-tune pre-trained text/image-to-video diffusion models on multi-view datasets of \textit{static} scenes (e.g., RealEstate10K \cite{zhou2018stereo}, ScanNet \cite{dai2017scannet}, DL3DV-10K \cite{ling2024dl3dv}) to integrate additional camera control signals.
They often rely on ControlNet-like hypernetworks \cite{zhang2023adding, wang2024motionctrl, he2024cameractrl, geng2024motion, gu2025diffusion}, raw camera extrinsics \cite{wang2024motionctrl}, camera ray coordinates \cite{he2024cameractrl, watson2024controlling}, Plücker coordinate embeddings \cite{kuang2025collaborative, bahmani2024ac3d, xu2024camco}, or combinations of multiple conditions \cite{popov2025camctrl3d}.

Since these methods derive multi-view priors primarily from \textit{static} scenes, they typically support camera-controlled generation from \textit{text} or \textit{image} inputs and are not designed to produce multiple videos that are mutually consistent.
In contrast, our work focuses on (\lowercase\expandafter{\romannumeral1}) enabling camera control over an input \textit{video} and, more importantly, (\lowercase\expandafter{\romannumeral2}) generating multiple videos that remain consistent with each other.
This is more challenging, as our framework must replicate the inherent dynamics of the input video, such as object motion, while ensuring consistency across the generated videos.

\section{Reangle-A-Video}
\vspace{-2pt}
In this section, we provide a comprehensive discussion of our framework.
Given an input video describing an arbitrary 4D scene, we aim to generate synchronized multi-view videos of the same scene without relying on multi-view generative priors.
Leveraging pre-trained latent image and video diffusion models (Sec. \ref{sec: method-preliminary}), our approach decomposes the dynamic 4D scene into view-specific appearance \textit{(starting image)} and view-invariant motion \textit{(image-to-video)}, addressing each component separately.
We first embed the scene's view-invariant motion into a pre-trained video diffusion model using our novel \textit{self-supervised training with data augmentation} strategy.
Initially, to capture diverse perspectives from a single monocular video, we repeatedly perform point-based warping to generate a set of warped videos (Sec. \ref{sec: method-warping}).
These videos, together with the original video, form the training dataset for fine-tuning a pre-trained image-to-video diffusion model with a masked diffusion objective (Sec. \ref{sec: method-training}).
To achieve \textit{(b) dynamic camera control}, we sample videos using the fine-tuned model with the original first frame as input.
In contrast, for \textit{(a) static view transport}, we generate view-transported starting images by inpainting the warped first frames under an inference-time view consistency guidance using an off-the-shelf multi-view stereo reconstruction network (Sec. \ref{sec: method-inpainting}).

\vspace{-2pt}
\subsection{Preliminary: Latent Diffusion Models}
\label{sec: method-preliminary}

Diffusion models \cite{sohl2015deep, ho2020denoising, song2020score} generate clean images or videos from Gaussian noise via an iterative denoising process.
This process reverses a fixed, time-dependent forward diffusion process, which gradually corrupts the data by adding Gaussian noise.
In latent diffusion models \cite{Rombach2021HighResolutionIS, blattmann2023align}, this process operates in the lower-dimensional latent space of a pre-trained VAE \cite{kingma2013auto}, comprised of an encoder $\mathcal{E}(\cdot)$ and a decoder $\mathcal{D}(\cdot)$.
Given a clean sample, $\xb_0 \sim p_{\text{data}}(\xb)$, it is first compressed to a latent representation $\zb_0 = \mathcal{E}(\xb_0)$.
Gaussian noise $\epsilonb \sim \mathcal{N}(0, I)$ is then added to produce intermediate noisy latents via the forward process $\zb_t = \alpha_t \zb_0 + \sigma_t \epsilonb,$
where $t$ denotes the diffusion timestep, and $\alpha_t$, $\sigma_t$ are noise scheduler parameters.
The training objective is to learn a denoiser network $\epsilonb_{\theta}$ that minimizes:
\vspace{-1mm}
\begin{equation}
    \label{eq: diffusion loss}
    \Eb_{\epsilonb \sim \mathcal{N}(0, I), \zb_t \sim p, t, c} \big[ \norm{\epsilonb - \epsilonb_{\theta} (\zb_t, t, c)}_2^2 \big],
    \vspace{-1mm}
\end{equation}
with condition $c$ provided by a text prompt, image, or both.
Once trained, the diffusion model generates clean samples by iteratively denoising a pure Gaussian noise.

\subsection{Stage \uppercase\expandafter{\romannumeral1}: Point-based Warping for Training Data Augmentation}
\label{sec: method-warping}
Given an input video $\xb^{1:N}$ of $N$ frames, our goal is to construct a training dataset for the subsequent fine-tuning stage, by lifting pixels into time-aware 3D point clouds and reprojecting them onto the image plane with target perspectives.

For each frame $\xb^i, i \in \{1, ..., N\}$, we first estimate its depth map $\mathbf{D}^i$ using a monocular depth estimator \cite{yang2024depth}.
The corresponding point cloud $\mathcal{P}^i$ is then generated from the RGBD image $[\xb^i, \mathbf{D}^i]$ as follows:
\vspace{-1mm}
\begin{equation}
    \label{eq: pixel-lifting}
    \mathcal{P}^i = \phi_{2\rightarrow3}([\xb^i, \mathbf{D}^i], \mathbf{K}, \mathbf{P}^i_\text{src}),
    \vspace{-1mm}
\end{equation}
where $\mathbf{K}$ is the camera intrinsic matrix, $\mathbf{P}^i_\text{src}$ is the extrinsic matrix for frame $\xb^i$, and $\phi_{2\rightarrow3}$ is the function that lifts the RGBD image into a point cloud.
Note that $\mathbf{K}$ and $\mathbf{P}^i_\text{src}$ are set by convention as in \cite{chung2023luciddreamer}, since they are intractable in open-domain videos.

Next, we define $M$ target extrinsic matrix trajectories $\Phi_{1:M}$, where each trajectory is given by 
\vspace{-1mm}
\begin{equation}
    \label{eq: target camera poses}
    \Phi_j = \{ \mathbf{P}^1_j, ..., \mathbf{P}^{N}_j \}, \; j \in \{1, ..., M\}.
    \vspace{-1mm}
\end{equation}
Each extrinsic matrix $\mathbf{P}^i_j$ comprises a rotation matrix $\mathbf{R} \in \mathbb{R}^{3 \times 3}$ and a translation vector $\mathbf{t} \in \mathbb{R}^{3 \times 1}$, which together transform the point cloud $\mathcal{P}^i$ into the target camera coordinate system.
In the \textit{static view transport} setting, each target trajectory $\Phi_j$ is constant across all frames, i.e., $\mathbf{P}^{1}_j = \mathbf{P}^{2}_j = ... = \mathbf{P}^{N}_j$;
for \textit{dynamic camera control}, each frame’s camera pose $\mathbf{P}^{i}_j$ is determined by incrementally moving and rotating from the previous pose $\mathbf{P}^{i-1}_j$, with the first target pose set to the pose of the first input frame ($\mathbf{P}^{1}_j = \mathbf{P}^1_\text{src}$).

Finally, we reproject each point cloud $\mathcal{P}^i$ to the image plane under the target perspective using $\mathbf{K}$ and $\mathbf{P}^{i}_j$ via the function $\phi_{3\rightarrow2}$:
\vspace{-1mm}
\begin{equation}
    \label{eq: reprojection}
   ( \hat{\xb}^i_j, \mb^i_j) = \phi_{3\rightarrow2}(\mathcal{P}^i, \textbf{K}, \textbf{P}^{i}_j),
    \vspace{-1mm}
\end{equation}
where $\hat{\xb}^i_j$ is the rendered warped image and $\mb^i_j$ is the corresponding visibility mask (1 for visible surfaces, 0 for invisible regions).
Repeating this process constructs our training dataset $\Omega$, which consists of $M$ pairs of warped videos and corresponding visibility masks: $\Omega  = \{ (\hat{\xb}^{1:N}_j, \mb^{1:N}_j) \; | \; j = 1, ..., M \}$.
We also add the input video $\xb^{1:N}$ with uniform visibility masks (all pixels set to 1), resulting in a total of $M{+}1$ video-mask pairs.

\subsection{Stage \uppercase\expandafter{\romannumeral2}: Multi-View Motion Learning}
\label{sec: method-training}
In this stage, we fine-tune a pre-trained image-to-video diffusion transformer to learn view-invariant motion
from the dataset $\Omega$ that captures diverse scene perspectives and corresponding pixels.
Here, ``motion" refers to the nuanced dynamics of each object—including its type, direction, and speed—as well as any inherent camera movement when the original video’s camera pose is not fixed.

To achieve lightweight fine-tuning while preserving the original model's prior, we employ Low-Rank Adaptation (LoRA) \cite{hu2022lora, ryu2023low}.
LoRA augments an attention layer by adding a residual path comprised of two low-rank matrices, $\theta_B \in \mathbb{R}^{d \times r}$, $\theta_A \in \mathbb{R}^{r \times k}$, to the original weight $\theta_0 \in \mathbb{R}^{d \times k}, r \ll \text{min}(d,k)$.
The modified forward pass is defined as:
\vspace{-1mm}
\begin{equation}
    \label{eq: lora}
    \theta = \theta_0 + \alpha \Delta \theta = \theta_0 + \alpha \theta_B \theta_A,
    \vspace{-1mm}
\end{equation}
where $\alpha$ controls the strength of the LoRA adjustment.
We implement this approach on a MM-DiT-based image-to-video diffusion model \cite{yang2024cogvideox}—which incorporates 3D full-attention in its MM-DiT blocks—by attaching LoRA layers to these 3D full-attention layers.

We then randomly sample a video $\xb^{1:N}$ and its corresponding visibility mask $\mb^{1:N}$ from the dataset $\Omega$ prepared in Sec. \ref{sec: method-warping}. Note that if $\xb^{1:N}$ is the original input video, its mask $\mb^{1:N}$ is uniformly filled with 1.
The video is compressed into latent space $\zb^{1:N}_0 = \mathcal{E}(\xb^{1:N})$, and the mask is downsampled to ${\mb}^{1:N}_\text{down}$ using spatio-temporal downsampling (see Sup. Sec.\ref{sup: 3d downsampling masks}).
The video latents are then noised, patchified, and unfolded into a long sequence to form video tokens. 
These, along with the text tokens, are fed into the video denoiser $\epsilonb_{\theta}$.
Since most sampled videos are warped and contain significant black regions, naive LoRA fine-tuning with the standard diffusion loss (Eq. (\ref{eq: diffusion loss})) inevitably degrades the original model prior and makes the model generate warped videos (see Supp. Sec. \ref{sup: masked diffusion loss}).
Thus, leveraging the compositionality of diffusion objectives \cite{gao2023mdtv2, zheng2023fast, avrahami2023break, zhang2024recapture, chou2024generating}, we fine-tune the LoRA layers $\Delta\theta$ using a masked diffusion loss that excludes invisible regions in it's loss computation:
\vspace{-1mm}
\begin{equation}
    \label{eq: masked diffusion loss}
    \Eb_{\epsilonb, \zb_t^{1:N}, {\mb}^{1:N}_\text{down}, t, c} \big[ \norm{ \epsilonb \odot {\mb}^{1:N}_\text{down} - \epsilonb_{\theta} (\zb_t^{1:N}, t, c) \odot {\mb}^{1:N}_\text{down} }_2^2 \big],
\end{equation}
where condition $c$ includes both the input text and image, with the latter being the first frame of the sampled video.
\footnote{
Following \cite{yang2024cogvideox}, the image input is encoded into latent space, then concatenated along the channel dimension with the noisy video latents.
}

For text input, \textit{(b) dynamic camera control} necessitates explicit specification of the camera movement type. 
As shown in Fig. \ref{fig: method-motion-learning}-\textit{(b)}, during the dynamic camera control training, the image-to-video model always starts with the same first frame—the original video's first frame—which is uniformly used across warped videos (see Sec. \ref{sec: method-warping}).
As a result, the model cannot infer the desired camera movement from the uniform starting frame alone.
To resolve this, we explicitly include the camera movement type in the text input during both training and inference.
Our LoRA layers, attached to the 3D full-attention modules, enable interactions between text and video tokens, learning to distinguish between different camera movements and map the corresponding text tokens to the appropriate video tokens.

The proposed training pipelines are outlined in Fig. \ref{fig: method-motion-learning}.
This synchronized few-shot training strategy prevents the video model from overfitting to a specific view. 
Instead, it learns general scene motion, enabling video generation even from viewpoints that were unseen during the training (see Sec. \ref{sec: ablation}), while ensuring that all output videos are synchronized (e.g., with consistent object motion speeds).

Once training is complete, for the \textit{(b) dynamic camera control} setup, we sample videos directly using the original video's first frame as the input image and text specifying the desired camera movement.
However, to achieve \textit{(a) static view transport}, we require a starting input image that captures the scene from the desired viewpoint—a process addressed in the next stage.

\begin{figure}[!t]
    \centering
    \includegraphics[width=\columnwidth]
    {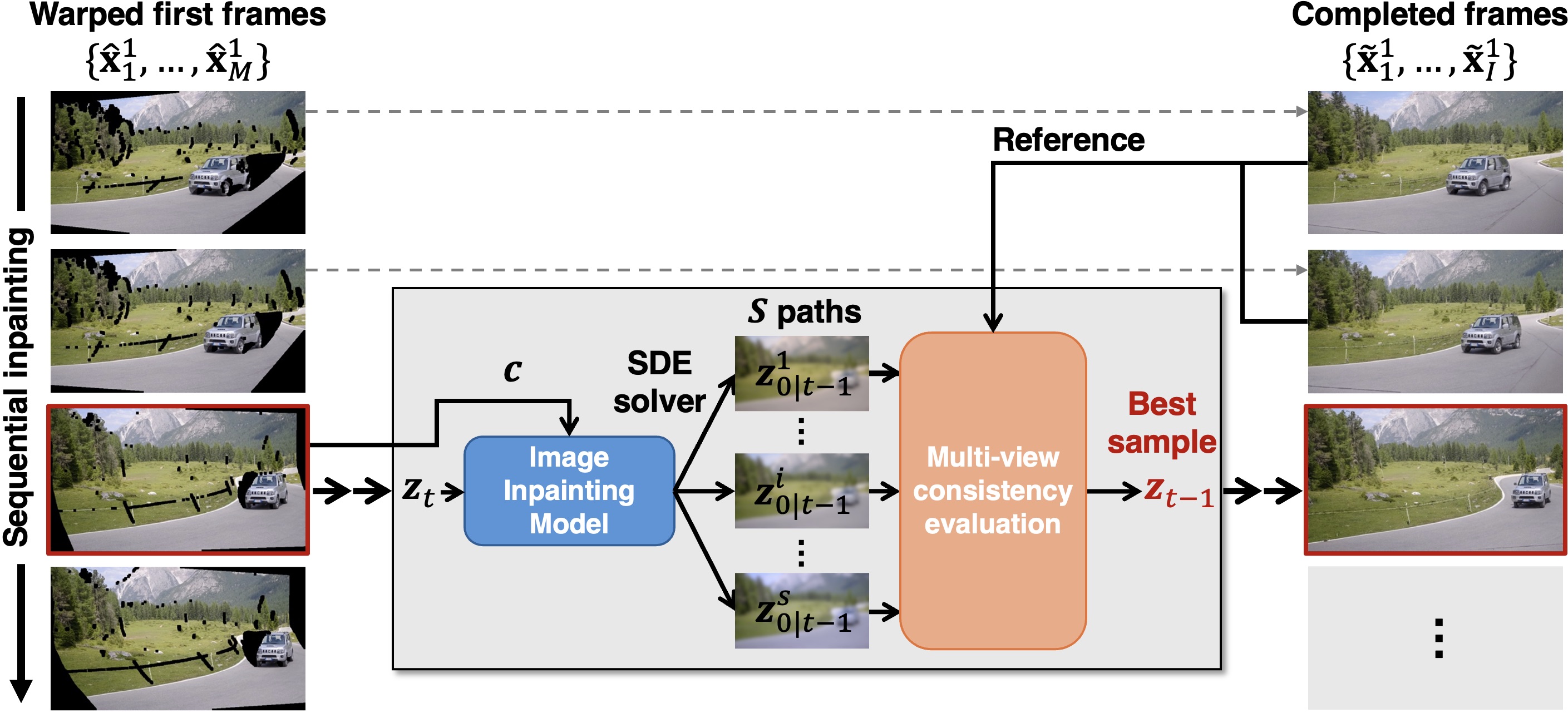}
    \caption{
    \textbf{Multi-view consistent image inpainting using stochastic control guidance.}
    In experiments, we set $S=25$.
    }
    \vspace{-2mm}
    \label{fig: multi-view-inpainting}
\end{figure}

\begin{figure}[!t]
    \centering
    \includegraphics[width=\columnwidth, height=7.4cm]
    {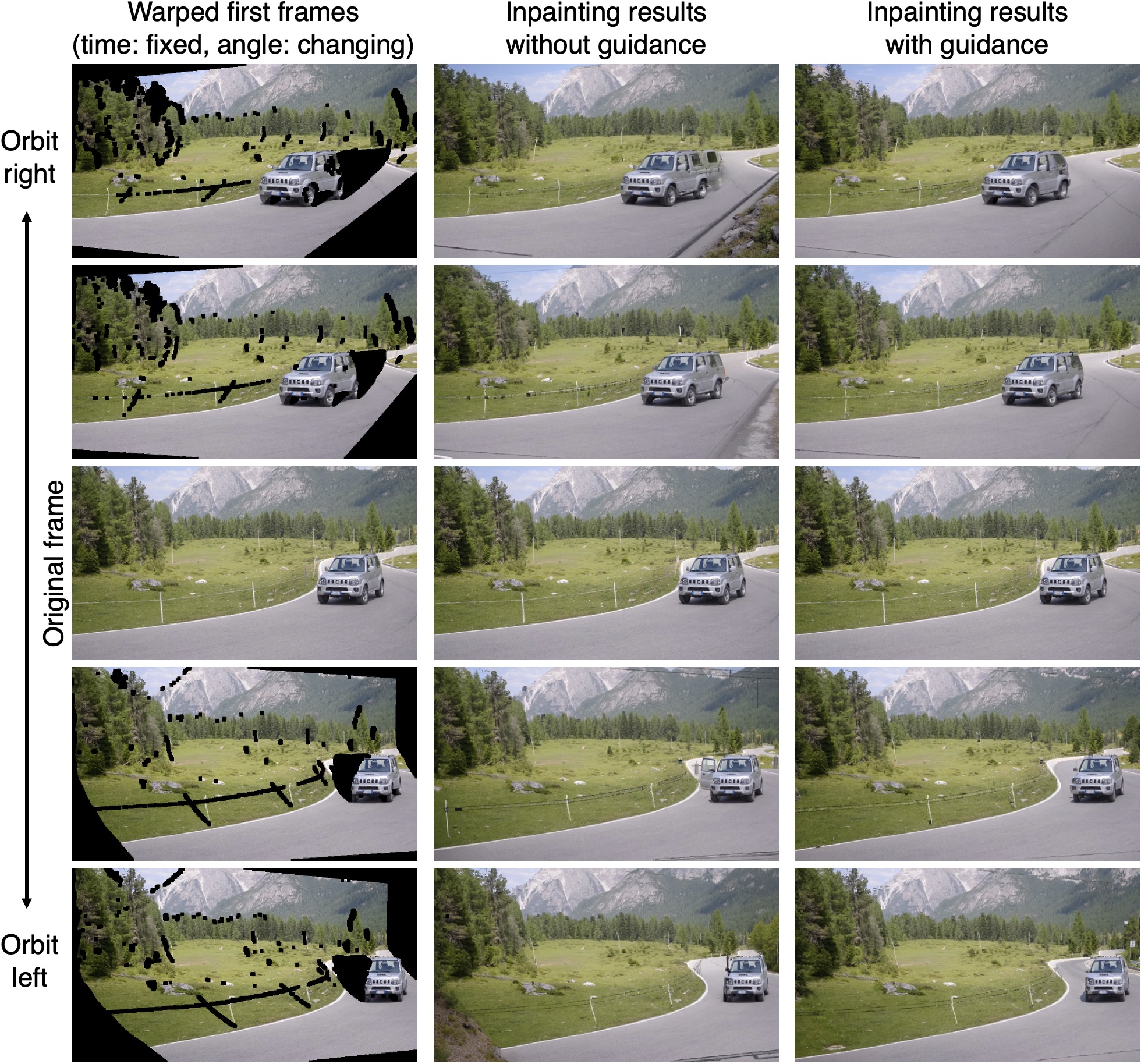}
    \caption{
    \textbf{Qualitative inpainting comparisons.}
    We compare naive inpainting to inpainting with stochastic control guidance.
    }
    \vspace{-3.5mm}
    \label{fig: inpainting-qualitative}
\end{figure}

\subsection{Stage \uppercase\expandafter{\romannumeral3}: Multi-View Image Inpainting}
\label{sec: method-inpainting}

We achieve multi-view consistent starting image generation using a warp-and-inpaint approach.
In Stage \uppercase\expandafter{\romannumeral1}, given an original image $\xb^1$ (the first frame of the input video), we render a set of warped images $\hat{\xb}^1_{1:M}$ from $M$ target viewpoints, along with corresponding binary masks $\mb^1_{1:M}$ that indicate invisible surfaces.
These missing regions are then filled via inpainting using image diffusion prior.
While state-of-the-art image generators \cite{flux, chen2024pixartdelta, esser2024scaling} yield plausible inpainting results on single warped images, independently inpainting each warped view fails to ensure cross-view consistency (see Fig. \ref{fig: inpainting-qualitative}, 2nd column).
Motivated by inference-time scaling practices \cite{huang2024symbolic, wallace2023end, yeh2024training, kim2024free}, we address the emerging inconsistencies by introducing a stochastic control guidance that enforces multi-view consistency during the denoising process using the DUSt3R multi-view stereo reconstruction model \cite{wang2024dust3r}.
Specifically, we sequentially inpaint warped images by leveraging previously inpainted views for consistency, guided by stochastic control guidance.
At each denoising step, we generate $S$ clean estimates from $S$ stochastic paths and compute a multi-view consistency score for each estimate with respect to the previously inpainted images $\{\Tilde{\xb}^1_1, ..., \Tilde{\xb}^1_I\}$.
Following \cite{asim2025met3r}, the score is calculated by projecting the images into a shared view using DUSt3R \cite{wang2024dust3r} and then computing the DINO feature similarity \cite{caron2021emerging} on the projected features.
We select the path with the highest score, add new noise to generate $S$ new paths, and iterate until the inpainting is complete
\footnote{
The process is repeated sequentially for all views, with the first two inpainted concurrently. To ensure initial consistency, we generate $S$ candidate versions per view at every sampling step, evaluate multi-view consistency for all $S^2$ pairs, and select the optimal pair for the next timestep. 
}.
We present detailed algorithms in Supp. Alg. \ref{supp: alg-diffusion} and \ref{supp: alg-flow}.

\begin{figure*}[!t]
    \centering
    \includegraphics[width=\textwidth]
    {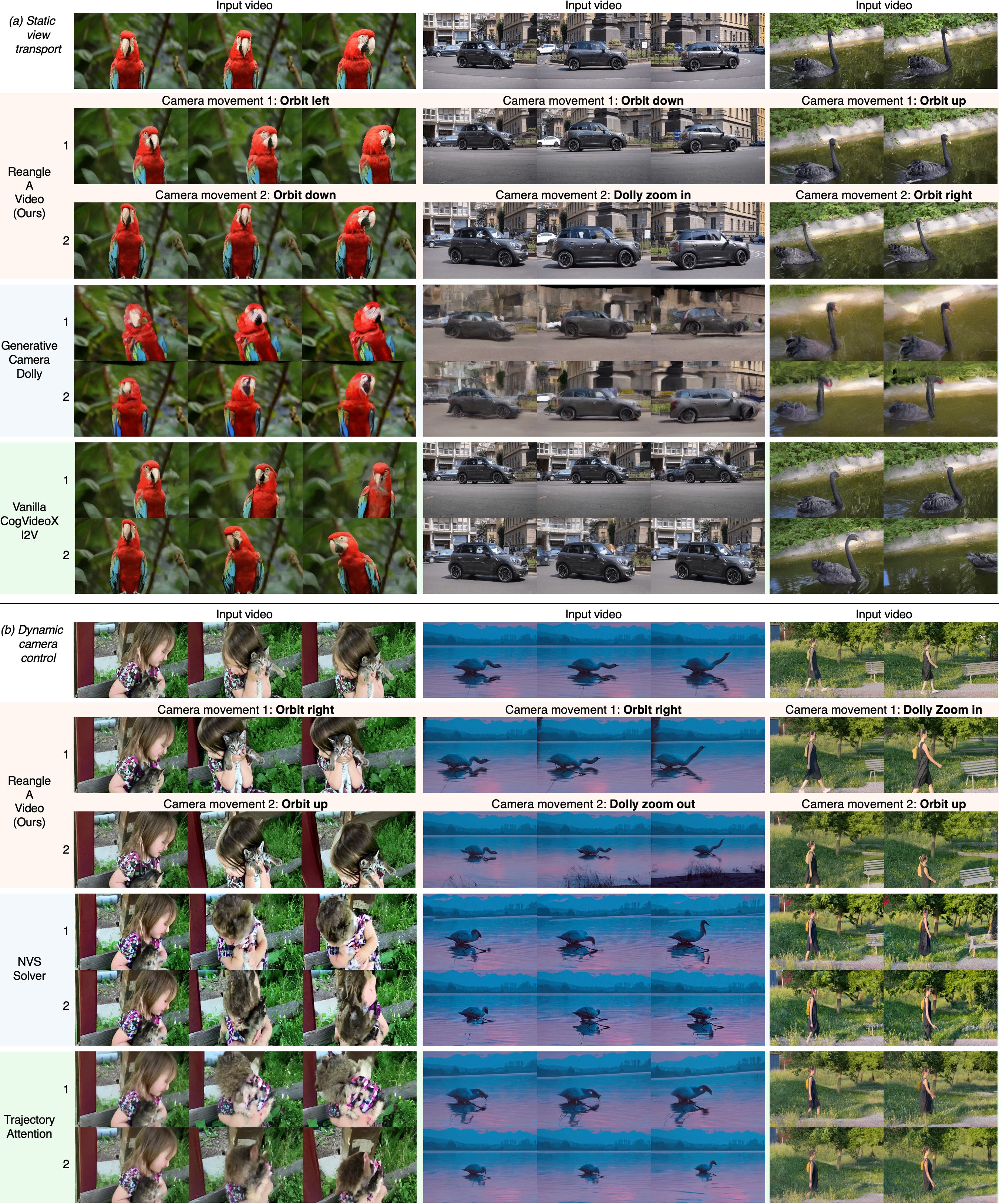}
    \caption{
    \textbf{Qualitative comparisons.}
    Top half shows \textit{(a) Static view transport} and bottom half presents \textit{(b) Dynamic camera control} results.
    The first row in each half displays the input videos, and for each input video, two generated videos corresponding to target cameras (1 and 2) are shown for each method.
    Across baseline, same camera parameters were used for each 1,2.
    Visit \href{https://hyeonho99.github.io/reangle-a-video/}{our page} for full-video results.
    }
    \label{fig: qualitative-results}
\end{figure*}

\begin{table*}[!t]
\centering
\newcommand{\first}[1]{\textbf{#1}}
\newcommand{\second}[1]{\underline{#1}}
\definecolor{Gray}{gray}{0.9}
\caption{
      \textbf{Quantitative comparison results}. 
      Reangle-A-Video is evaluated in two modes—\textbf{(a) Static View Transport} and \textbf{(b) Dynamic Camera Control}—against the baselines.
      VBench metrics \cite{huang2024vbench++} are presented on the left, other metrics \cite{asim2025met3r, heusel2017gans, unterthiner2018towards} appear on the right.
      }
  \begin{adjustbox}{max width=0.98\textwidth}
  \begin{tabular}{cccccccc|ccc}
    \toprule
       & Subject & Background & Temporal & Motion & Dynamic & Aesthetic & Imaging & \multirow{2}{*}{MEt3R$\downarrow$} & \multirow{2}{*}{FID$\downarrow$} & \multirow{2}{*}{FVD$\downarrow$} \\
       \textbf{(a) Static View Transport} & Consistency & Consistency & Flickering & Smoothness & Degree & Quality & Quality &   &  \\
    \hline
   Generative Camera Dolly \cite{van2024generative} & 0.8849 & 0.9107 & 0.8731 & 0.9119 & 0.7608 & 0.4052 & 0.5524 & 0.1237 & 155.15 & 5264.7 \\
   Vanilla CogVideoX \cite{yang2024cogvideox} & 0.9448 & \first{0.9398} & 0.9738 & 0.9859 & 0.7286 & 0.4997 & \first{0.6475} & 0.0539 & 79.621 & 3664.2 \\
   \rowcolor{Gray}
   Reangle-A-Video (Ours) & \first{0.9516} & 0.9327 & \first{0.9764} & \first{0.9907} & \first{0.7657} & \first{0.5160} & 0.6414 & \textbf{0.0412} & \textbf{53.448} & \textbf{2690.9} \\
   \bottomrule
   \textbf{(b) Dynamic Camera Control} & & & & & & & & & \\
   \hline
   NVS-Solver \cite{you2024nvs}  & 0.9037 & 0.9325 & 0.9049 & 0.9521 & 0.8809 & 0.5023 & \first{0.6411} & 0.1090 & 95.815 & 3516.5 \\
   Trajectory Attention \cite{xiao2024trajectory} & 0.8984 & 0.9288 & 0.9342 & 0.9658 & \first{0.8889} & 0.4854 & 0.5990 & 0.0965 & 109.20 & 3624.9 \\
   \rowcolor{Gray}
   Reangle-A-Video (Ours) & \first{0.9140} & \first{0.9364} & \first{0.9386} & \first{0.9794} & 0.8884 & \first{0.5238} & 0.6271 & \textbf{0.0648} & \textbf{74.194} & \textbf{3019.7} \\
    \bottomrule
    \end{tabular}
    \end{adjustbox}
    \vspace{-2.5mm}
  \label{tab: quantitative-result}
\end{table*}

\section{Experiments}
\vspace{-0.5mm}

\subsection{Implementation Details}
\label{sec: impldetail}
We experiment with $28$ publicly sourced videos \cite{pont20172017, pexels}, covering diverse scenes with varying objects, scene motions, and physical environments.
Each video consists of 49 frames at a resolution of $480 {\times} 720$.
On average, we generate videos from 3.5 different camera viewpoints/movements per input, resulting in a total of $98$ generated videos.

For \textbf{Stage \uppercase\expandafter{\romannumeral1}}, Depth Anything V2 model \cite{yang2024depth} is used to estimate the depth maps of the input videos.
For \textbf{Stage \uppercase\expandafter{\romannumeral2}} and \textbf{\uppercase\expandafter{\romannumeral3}}, we use the CogVideoX-5b image-to-video diffusion model \cite{yang2024cogvideox}, and FLUX text-to-image model \cite{flux} with a pre-trained inpainting ControlNet \cite{inpaint-controlnet} attached to it.

Upon video model fine-tuning, we set the LoRA rank to $128$ and optimize the LoRA layers over $400$ steps, where the fine-tuned parameters account for only about 2\% of the original Video DiT parameters 
\footnote{To enable video model fine-tuning within a 40GB VRAM constraint, we employ gradient checkpointing, which reduces memory usage at the expense of slower gradient back-propagation.}.
We use AdamW \cite{loshchilov2017decoupled} optimizer with a learning rate of $1e{-}4$ and a weight decay of $1e{-}3$.
The fine-tuning takes about an hour.
For the \textit{static view transport} mode, we set $M=12$ warped videos; for \textit{dynamic camera control}, we set $M=6$ (see Supp. Sec. \ref{sup: training dataset composition}).
For video sampling, we apply $40$ sampling steps with a CFG \cite{ho2022classifier} scale of $6.0$.
During multi-view image inpainting, we first resize the input to $1024 \times 1024$ and then restore it to its original resolution.
We use MEt3R~\cite{asim2025met3r} as the reward function for stochastic control with $S=25$. 
To introduce stochasticity for the path control, we employ an SDE-based sampler with 50 steps, ensuring the same marginal distribution as the original ODE of the flow trajectory (see Supp. Sec. \ref{supp: sec: inpainting details}).
All experiments, including inference and fine-tuning, are performed using 40GB A100 GPUs.

\vspace{-2mm}
\subsection{Comparisons}
\vspace{-1mm}
\subsubsection{Baselines}
\vspace{-1mm}
Multi-view/camera synchronized video generation from an input video remains largely underexplored. 
For \textit{(a) Static view transport}, the closest works include \textbf{Generative Camera Dolly} \cite{van2024generative} and the closed-source multi-view video foundation model CAT4D \cite{wu2024cat4d}.
Although GS-DiT \cite{bian2025gs} aligns with our framework, its code was unavailable during our research.
In addition to GCD, we evaluate a baseline using naive CogVideoX-I2V inference (\textbf{Vanilla CogVideoX})\cite{yang2024cogvideox}, which employs the same input frame as our approach.
For \textit{(b) Dynamic camera control} over videos, we compare against two state-of-the-art methods, \textbf{NVS-Solver} \cite{you2024nvs} and \textbf{Trajectory Attention} \cite{xiao2024trajectory}. 
While Recapture \cite{zhang2024recapture} is closely related, its code is not available, and it employs the proprietary multi-view image foundation model CAT3D \cite{gao2024cat3d}.

\vspace{-1mm}
\subsubsection{Qualitative Results}
\vspace{-1mm}
Comprehensive qualitative results are shown in Fig. \ref{fig: qualitative-results}. For \textit{(a) Static View Transport}, Generative Camera Dolly \cite{van2024generative} struggles with real-world videos (trained on synthetic data \cite{greff2022kubric}), while Vanilla CogVideoX \cite{yang2024cogvideox} fails to capture the input video's motion.
In contrast, Reangle-A-Video accurately reproduces the input motion from the target viewpoint.
For \textit{(b) Dynamic Camera Control}, NVS-Solver \cite{you2024nvs} and Trajectory Attention \cite{xiao2024trajectory} either confuse foreground objects (e.g., a girl's and a cat's head in the first video), fail to capture precise motion (e.g., neck movement in the second video), or miss background elements (e.g., a bench in the third video).
Our method faithfully regenerates the input video's motion, preserves object appearance, and accurately follows the target camera movement.

\vspace{-1mm}
\subsubsection{Quantitative Results}
\vspace{-1mm}
\textbf{Automatic metrics.}
For automated evaluation, we first use VBench \cite{huang2024vbench++}, which assesses generated videos across various disentangled dimensions.
As shown in Tab. \ref{tab: quantitative-result}-left, our method outperforms baselines in most metrics for both \textit{static view transport} and \textit{dynamic camera control} modes.
Notably, compared to vanilla CogVideoX I2V—which uses the same input image—our approach maintain robust performance even when fine-tuned with warped videos.
Additionally, Tab. \ref{tab: quantitative-result}-right presents evaluation results using FID \cite{heusel2017gans}, FVD \cite{unterthiner2018towards}, and the recently proposed multi-view consistency metric MEt3R \cite{asim2025met3r}.

\begin{figure}[!t]
    \centering
    \includegraphics[width=\columnwidth]
    {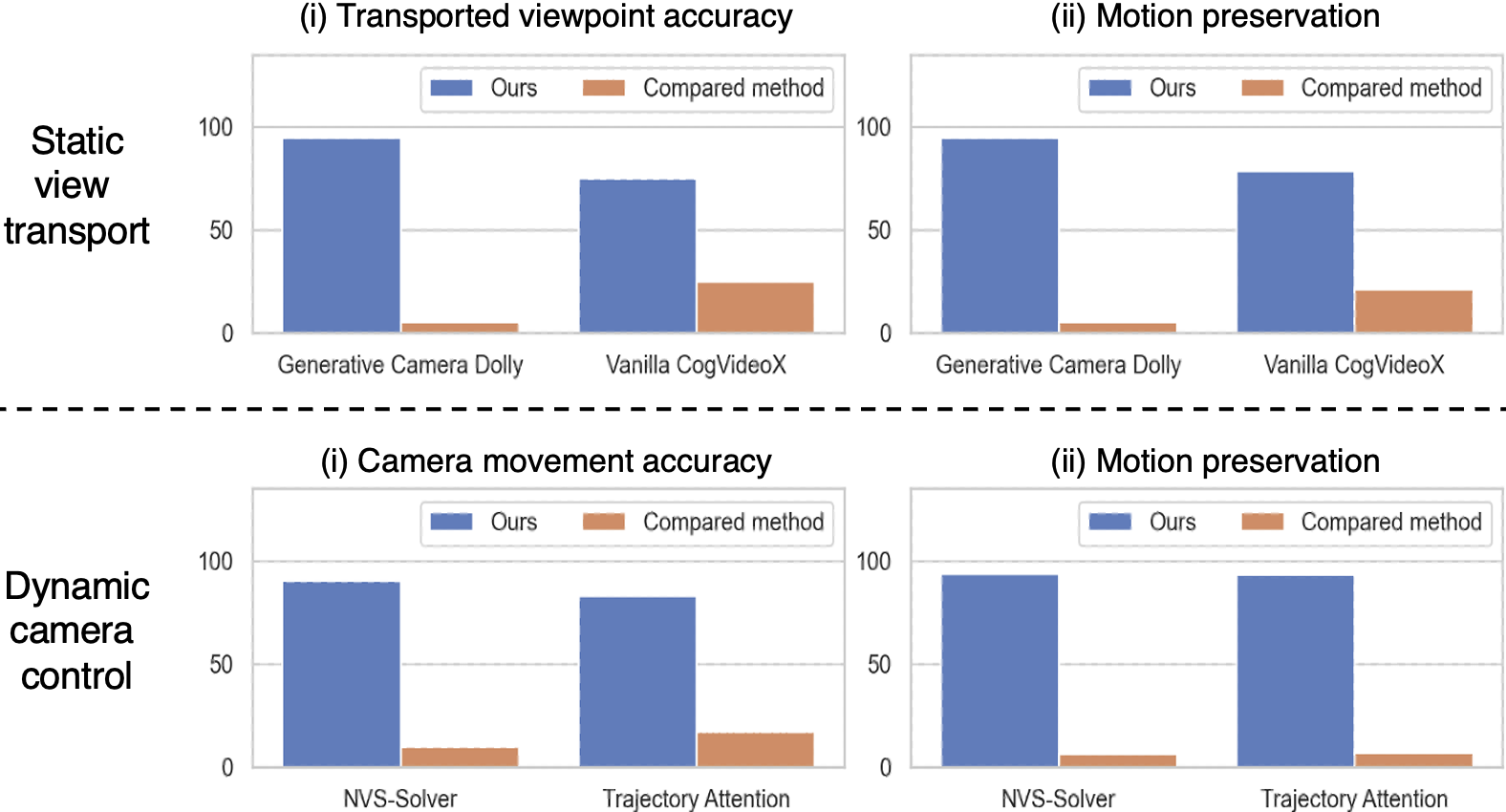}
    \caption{
    \textbf{User study results.}
    Top: Static view transport results.
    Bottom: Dynamic camera control results.
    }
    \label{fig: user-study-results}
\end{figure}

\noindent\textbf{Human evaluation.}
We further assess Reangle-A-Video against baselines via a user study involving 36 participants who compare our results with randomly selected baselines.
For the \textit{static view transport} evaluation, participants rate:
(\lowercase\expandafter{\romannumeral1}) accuracy of the transported viewpoint,
(\lowercase\expandafter{\romannumeral2}) preservation of the input video's motion.
For the \textit{dynamic camera control} evaluation, participants assess:
(\lowercase\expandafter{\romannumeral1}) the accuracy of the target camera movement,
(\lowercase\expandafter{\romannumeral2}) the preservation of the input video's motion in the output.
As reported in Fig. \ref{fig: user-study-results}, our method outperforms the baselines in all aspects.

\begin{table}[!t]
\centering
\newcommand{\first}[1]{\textbf{#1}}
\newcommand{\second}[1]{\underline{#1}}
\definecolor{Gray}{gray}{0.9}
\caption{
    Quantitative evaluation of multi-view consistency in image inpainting with and without stochastic control guidance.
      }
  \begin{adjustbox}{max width=0.92\columnwidth}
  \begin{tabular}{cccc}
    \toprule
    Warped image inpainting    & MEt3R $\downarrow$ & SED $\downarrow$ & TSED $\uparrow$ \\
    \hline
   w/o stochastic control guidance  & 0.1431 & 1.1966 & 0.5241 \\
   \rowcolor{Gray}
   w/ stochastic control guidance & \textbf{0.1184} & \textbf{1.1844} & \textbf{0.5588} \\
    \bottomrule
    \end{tabular}
    \end{adjustbox}
    \vspace{-3mm}
  \label{tab: ablation-inpainting}
\end{table}

\subsection{Ablation Studies}
\label{sec: ablation}
\vspace{-1mm}

\begin{figure}[!t]
    \centering
    \includegraphics[width=\columnwidth]
    {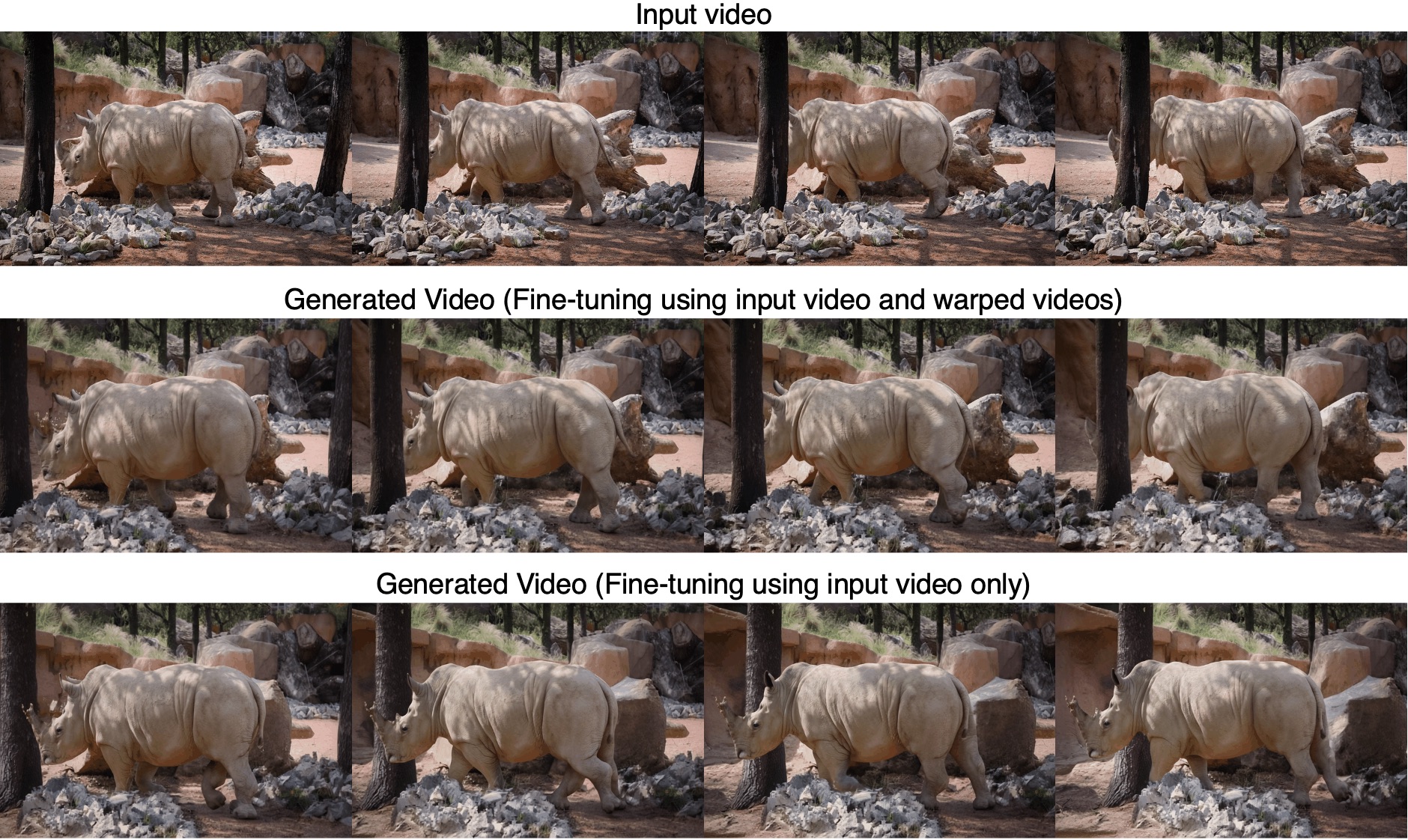}
    \caption{
    Novel view video generation with and without using warped video for training (target viewpoint: dolly zoom in).
    }
    \vspace{-1mm}
    \label{fig: ablate-warping-augmentation}
\end{figure}

\begin{figure}[!t]
    \centering
    \includegraphics[width=\columnwidth]
    {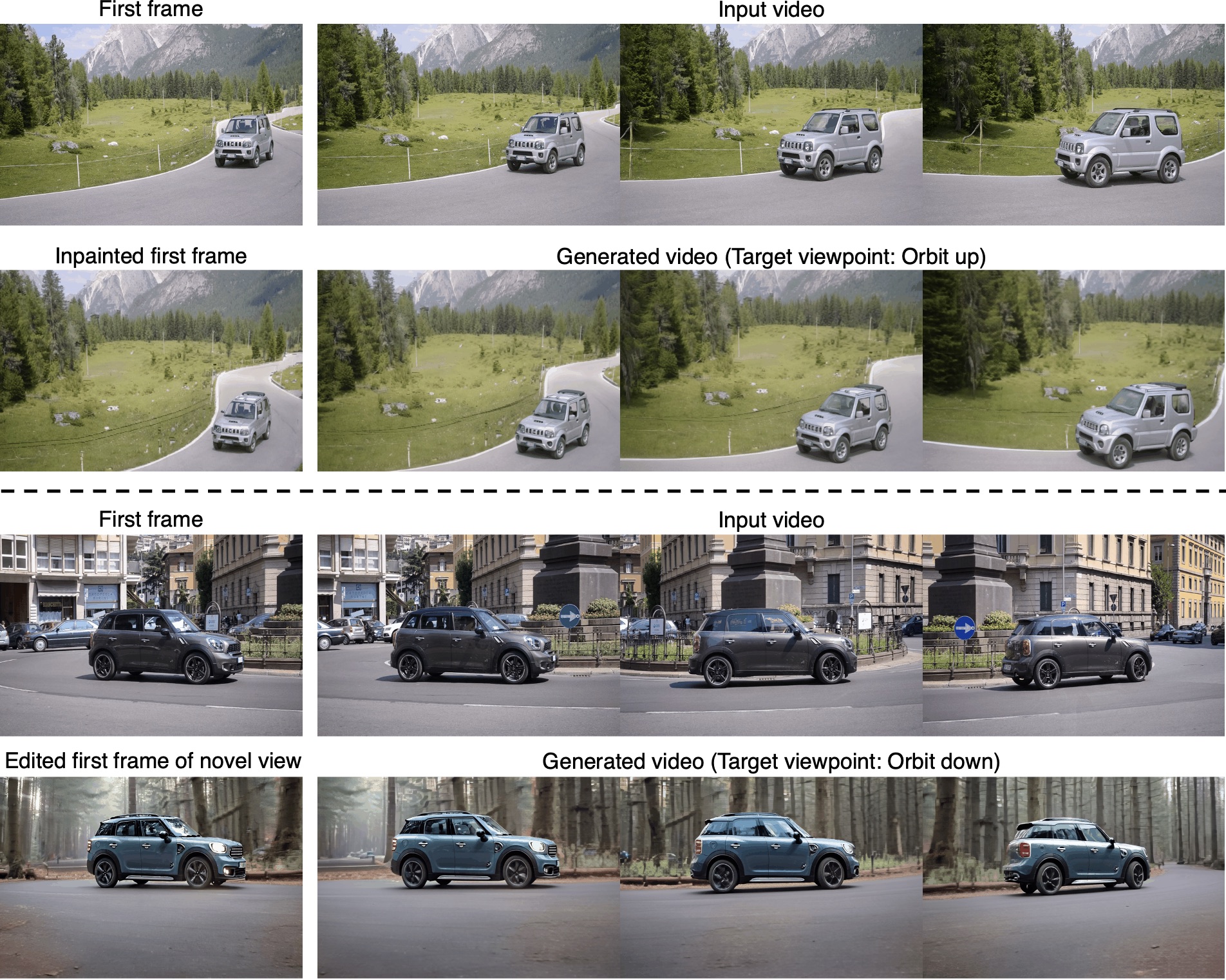}
    \caption{
    Top: Unseen view video generation. 
    Bottom: Novel view video generation using an appearance-edited first frame.
    }
    \vspace{-1mm}
    \label{fig: unseen-view_first-frame-edit}
\end{figure}

First, we ablate the stochastic control guidance by comparing it with naive inpainting (where each image is inpainted independently).
Tab. \ref{tab: ablation-inpainting} evaluates multi-view accuracy using the MEt3R~\cite{asim2025met3r}, SED, and TSED~\cite{yu2023long} metrics 
(with $T_e=1.25$ and $T_m=10$ for SED). 
Fig. \ref{fig: inpainting-qualitative} shows that our method fills target (invisible) regions consistently across multiple views.
Next, we ablate the necessity of our data augmentation strategy (i.e., the use of warped videos) in learning view-invariant motion.
Fig. \ref{fig: ablate-warping-augmentation} compares fine-tuning with only the original input video versus using both the original and warped videos.
The results indicate that relying solely on the original video fails to accurately capture motion—for example, the rhino moving in front of the tree. 
For quantitative assessment on input video's motion preservation, we employ user study (see Supp. Sec. \ref{sup: warped-videos-user-study}.)
In Fig. \ref{fig: unseen-view_first-frame-edit}-top, we demonstrate unseen view video generation. We exclude specific warped view videos (vertical up/down orbits) from the training dataset and fine-tune the video model without them. Then, using an inpainted first frame as input, we generate a video from that omitted view.
In Fig. \ref{fig: unseen-view_first-frame-edit}-bottom, we showcase novel view video generation guided by an edited first frame. Starting with an inpainted image representing the scene from a target viewpoint, we apply FlowEdit \cite{kulikov2024flowedit} to modify the image, then generate the novel-view video with our fine-tuned model.
In both cases, the generated videos faithfully follow the input video's motion, demonstrating that our few-shot training strategy is robust in both view and appearance.

\section{Conclusion}
\vspace{-1mm}
We have presented an approximate yet effective solution for generating synchronized multi-view videos from a single monocular real-world video—without relying on any multi-view generative priors.
Our approach enables an image-to-video diffusion model to learn view-invariant scene motion through self-supervised fine-tuning with video augmentation.
Then, we sample videos from the fine-tuned model using a set of multi-view consistent first-frame images, where the images are generated via a warp-and-inpaint process that enforces multi-view consistency during inference-time.

\paragraph{Limitations and future work.}
Input image quality is crucial in image-to-video generation, as the output cannot exceed the first frame's quality.
While our approach ensures \textit{multi-view consistency} in the warp-and-inpaint scheme, the warping stage is inherently prone to artifacts caused by camera and depth errors (Supp. Sec. \ref{sup: limitations}).
We believe that improved depth models could enhance both the warped and inpainted images.
Another limitation of our work is the need for scene-specific model tuning to achieve 4D video synthesis.
A promising future direction is to extend existing video datasets with warped videos to train 4D foundation models.

\section*{Acknowledgments}
We would like to thank Geon Yeong Park and Mingi Kwon for their valuable discussions at the early stage of the project.
This work was supported by the National Research Foundation of Korea under Grant
RS-2024-00336454, and by the Institute of Information \& Communications Technology Planning \& Evaluation (IITP) grant funded by the Korean government~(MSIT) (No. RS-2024-00457882, AI Research Hub Project; No. RS-2025-02304967, AI Star Fellowship~(KAIST)).

{
    \small
    \bibliographystyle{ieeenat_fullname}
    \bibliography{main}

\begin{thebibliography}{97}
\providecommand{\natexlab}[1]{#1}
\providecommand{\url}[1]{\texttt{#1}}
\expandafter\ifx\csname urlstyle\endcsname\relax
  \providecommand{\doi}[1]{doi: #1}\else
  \providecommand{\doi}{doi: \begingroup \urlstyle{rm}\Url}\fi

\bibitem[Asim et~al.(2025)Asim, Wewer, Wimmer, Schiele, and Lenssen]{asim2025met3r}
Mohammad Asim, Christopher Wewer, Thomas Wimmer, Bernt Schiele, and Jan~Eric Lenssen.
\newblock Met3r: Measuring multi-view consistency in generated images.
\newblock \emph{arXiv preprint arXiv:2501.06336}, 2025.

\bibitem[Avrahami et~al.(2023)Avrahami, Aberman, Fried, Cohen-Or, and Lischinski]{avrahami2023break}
Omri Avrahami, Kfir Aberman, Ohad Fried, Daniel Cohen-Or, and Dani Lischinski.
\newblock Break-a-scene: Extracting multiple concepts from a single image.
\newblock In \emph{SIGGRAPH Asia 2023 Conference Papers}, pages 1--12, 2023.

\bibitem[Bahmani et~al.(2024)Bahmani, Skorokhodov, Qian, Siarohin, Menapace, Tagliasacchi, Lindell, and Tulyakov]{bahmani2024ac3d}
Sherwin Bahmani, Ivan Skorokhodov, Guocheng Qian, Aliaksandr Siarohin, Willi Menapace, Andrea Tagliasacchi, David~B Lindell, and Sergey Tulyakov.
\newblock Ac3d: Analyzing and improving 3d camera control in video diffusion transformers.
\newblock \emph{arXiv preprint arXiv:2411.18673}, 2024.

\bibitem[Bai et~al.(2024)Bai, Xia, Wang, Yuan, Fu, Liu, Hu, Wan, and Zhang]{bai2024syncammaster}
Jianhong Bai, Menghan Xia, Xintao Wang, Ziyang Yuan, Xiao Fu, Zuozhu Liu, Haoji Hu, Pengfei Wan, and Di Zhang.
\newblock Syncammaster: Synchronizing multi-camera video generation from diverse viewpoints.
\newblock \emph{arXiv preprint arXiv:2412.07760}, 2024.

\bibitem[Bar-Tal et~al.(2024)Bar-Tal, Chefer, Tov, Herrmann, Paiss, Zada, Ephrat, Hur, Liu, Raj, et~al.]{bar2024lumiere}
Omer Bar-Tal, Hila Chefer, Omer Tov, Charles Herrmann, Roni Paiss, Shiran Zada, Ariel Ephrat, Junhwa Hur, Guanghui Liu, Amit Raj, et~al.
\newblock Lumiere: A space-time diffusion model for video generation.
\newblock In \emph{SIGGRAPH Asia 2024 Conference Papers}, pages 1--11, 2024.

\bibitem[Bian et~al.(2025)Bian, Huang, Shi, Li, Wang, and Li]{bian2025gs}
Weikang Bian, Zhaoyang Huang, Xiaoyu Shi, Yijin Li, Fu-Yun Wang, and Hongsheng Li.
\newblock Gs-dit: Advancing video generation with pseudo 4d gaussian fields through efficient dense 3d point tracking.
\newblock \emph{arXiv preprint arXiv:2501.02690}, 2025.

\bibitem[Blattmann et~al.(2023{\natexlab{a}})Blattmann, Dockhorn, Kulal, Mendelevitch, Kilian, Lorenz, Levi, English, Voleti, Letts, et~al.]{blattmann2023stable}
Andreas Blattmann, Tim Dockhorn, Sumith Kulal, Daniel Mendelevitch, Maciej Kilian, Dominik Lorenz, Yam Levi, Zion English, Vikram Voleti, Adam Letts, et~al.
\newblock Stable video diffusion: Scaling latent video diffusion models to large datasets.
\newblock \emph{arXiv preprint arXiv:2311.15127}, 2023{\natexlab{a}}.

\bibitem[Blattmann et~al.(2023{\natexlab{b}})Blattmann, Rombach, Ling, Dockhorn, Kim, Fidler, and Kreis]{blattmann2023align}
Andreas Blattmann, Robin Rombach, Huan Ling, Tim Dockhorn, Seung~Wook Kim, Sanja Fidler, and Karsten Kreis.
\newblock Align your latents: High-resolution video synthesis with latent diffusion models.
\newblock In \emph{Proceedings of the IEEE/CVF Conference on Computer Vision and Pattern Recognition}, pages 22563--22575, 2023{\natexlab{b}}.

\bibitem[Brooks et~al.(2024)Brooks, Peebles, Holmes, DePue, Guo, Jing, Schnurr, Taylor, Luhman, Luhman, Ng, Wang, and Ramesh]{videoworldsimulators2024}
Tim Brooks, Bill Peebles, Connor Holmes, Will DePue, Yufei Guo, Li Jing, David Schnurr, Joe Taylor, Troy Luhman, Eric Luhman, Clarence Ng, Ricky Wang, and Aditya Ramesh.
\newblock Video generation models as world simulators.
\newblock 2024.

\bibitem[Burgert et~al.(2025)Burgert, Xu, Xian, Pilarski, Clausen, He, Ma, Deng, Li, Mousavi, et~al.]{burgert2025go}
Ryan Burgert, Yuancheng Xu, Wenqi Xian, Oliver Pilarski, Pascal Clausen, Mingming He, Li Ma, Yitong Deng, Lingxiao Li, Mohsen Mousavi, et~al.
\newblock Go-with-the-flow: Motion-controllable video diffusion models using real-time warped noise.
\newblock \emph{arXiv preprint arXiv:2501.08331}, 2025.

\bibitem[Caron et~al.(2021)Caron, Touvron, Misra, J{\'e}gou, Mairal, Bojanowski, and Joulin]{caron2021emerging}
Mathilde Caron, Hugo Touvron, Ishan Misra, Herv{\'e} J{\'e}gou, Julien Mairal, Piotr Bojanowski, and Armand Joulin.
\newblock Emerging properties in self-supervised vision transformers.
\newblock In \emph{Proceedings of the IEEE/CVF international conference on computer vision}, pages 9650--9660, 2021.

\bibitem[Chefer et~al.(2024)Chefer, Zada, Paiss, Ephrat, Tov, Rubinstein, Wolf, Dekel, Michaeli, and Mosseri]{chefer2024still}
Hila Chefer, Shiran Zada, Roni Paiss, Ariel Ephrat, Omer Tov, Michael Rubinstein, Lior Wolf, Tali Dekel, Tomer Michaeli, and Inbar Mosseri.
\newblock Still-moving: Customized video generation without customized video data.
\newblock \emph{ACM Transactions on Graphics (TOG)}, 43\penalty0 (6):\penalty0 1--11, 2024.

\bibitem[Chen et~al.(2023)Chen, Yu, Ge, Yao, Xie, Wu, Wang, Kwok, Luo, Lu, et~al.]{chen2023pixart}
Junsong Chen, Jincheng Yu, Chongjian Ge, Lewei Yao, Enze Xie, Yue Wu, Zhongdao Wang, James Kwok, Ping Luo, Huchuan Lu, et~al.
\newblock Pixart-$\alpha$: Fast training of diffusion transformer for photorealistic text-to-image synthesis.
\newblock \emph{arXiv preprint arXiv:2310.00426}, 2023.

\bibitem[Chen et~al.(2024{\natexlab{a}})Chen, Ge, Xie, Wu, Yao, Ren, Wang, Luo, Lu, and Li]{chen2024pixart}
Junsong Chen, Chongjian Ge, Enze Xie, Yue Wu, Lewei Yao, Xiaozhe Ren, Zhongdao Wang, Ping Luo, Huchuan Lu, and Zhenguo Li.
\newblock Pixart-$\sigma$: Weak-to-strong training of diffusion transformer for 4k text-to-image generation.
\newblock In \emph{European Conference on Computer Vision}, pages 74--91. Springer, 2024{\natexlab{a}}.

\bibitem[Chen et~al.(2024{\natexlab{b}})Chen, Wu, Luo, Xie, Paul, Luo, Zhao, and Li]{chen2024pixartdelta}
Junsong Chen, Yue Wu, Simian Luo, Enze Xie, Sayak Paul, Ping Luo, Hang Zhao, and Zhenguo Li.
\newblock Pixart-$\delta$: Fast and controllable image generation with latent consistency models.
\newblock \emph{arXiv preprint arXiv:2401.05252}, 2024{\natexlab{b}}.

\bibitem[Cheong et~al.(2024)Cheong, Ceylan, Mustafa, Gilbert, and Huang]{cheong2024boosting}
Soon~Yau Cheong, Duygu Ceylan, Armin Mustafa, Andrew Gilbert, and Chun-Hao~Paul Huang.
\newblock Boosting camera motion control for video diffusion transformers.
\newblock \emph{arXiv preprint arXiv:2410.10802}, 2024.

\bibitem[Chou et~al.(2024)Chou, Zhang, Bi, Tan, Xu, Luan, Hariharan, and Snavely]{chou2024generating}
Gene Chou, Kai Zhang, Sai Bi, Hao Tan, Zexiang Xu, Fujun Luan, Bharath Hariharan, and Noah Snavely.
\newblock Generating 3d-consistent videos from unposed internet photos.
\newblock \emph{arXiv preprint arXiv:2411.13549}, 2024.

\bibitem[Chung et~al.(2023)Chung, Lee, Nam, Lee, and Lee]{chung2023luciddreamer}
Jaeyoung Chung, Suyoung Lee, Hyeongjin Nam, Jaerin Lee, and Kyoung~Mu Lee.
\newblock Luciddreamer: Domain-free generation of 3d gaussian splatting scenes.
\newblock \emph{arXiv preprint arXiv:2311.13384}, 2023.

\bibitem[Creative and Team(2025)]{inpaint-controlnet}
Alimama~Smart Creative and AI~Application Team.
\newblock Flux inpainting controlnet, 2025.
\newblock URL https://huggingface.co/alimama-creative/FLUX.1-dev-Controlnet-Inpainting-Beta.

\bibitem[Dai et~al.(2017)Dai, Chang, Savva, Halber, Funkhouser, and Nie{\ss}ner]{dai2017scannet}
Angela Dai, Angel~X Chang, Manolis Savva, Maciej Halber, Thomas Funkhouser, and Matthias Nie{\ss}ner.
\newblock Scannet: Richly-annotated 3d reconstructions of indoor scenes.
\newblock In \emph{Proceedings of the IEEE conference on computer vision and pattern recognition}, pages 5828--5839, 2017.

\bibitem[Deitke et~al.(2023)Deitke, Schwenk, Salvador, Weihs, Michel, VanderBilt, Schmidt, Ehsani, Kembhavi, and Farhadi]{deitke2023objaverse}
Matt Deitke, Dustin Schwenk, Jordi Salvador, Luca Weihs, Oscar Michel, Eli VanderBilt, Ludwig Schmidt, Kiana Ehsani, Aniruddha Kembhavi, and Ali Farhadi.
\newblock Objaverse: A universe of annotated 3d objects.
\newblock In \emph{Proceedings of the IEEE/CVF Conference on Computer Vision and Pattern Recognition}, pages 13142--13153, 2023.

\bibitem[Deitke et~al.(2024)Deitke, Liu, Wallingford, Ngo, Michel, Kusupati, Fan, Laforte, Voleti, Gadre, et~al.]{deitke2024objaverse}
Matt Deitke, Ruoshi Liu, Matthew Wallingford, Huong Ngo, Oscar Michel, Aditya Kusupati, Alan Fan, Christian Laforte, Vikram Voleti, Samir~Yitzhak Gadre, et~al.
\newblock Objaverse-xl: A universe of 10m+ 3d objects.
\newblock \emph{Advances in Neural Information Processing Systems}, 36, 2024.

\bibitem[Esser et~al.(2024)Esser, Kulal, Blattmann, Entezari, M{\"u}ller, Saini, Levi, Lorenz, Sauer, Boesel, et~al.]{esser2024scaling}
Patrick Esser, Sumith Kulal, Andreas Blattmann, Rahim Entezari, Jonas M{\"u}ller, Harry Saini, Yam Levi, Dominik Lorenz, Axel Sauer, Frederic Boesel, et~al.
\newblock Scaling rectified flow transformers for high-resolution image synthesis.
\newblock In \emph{Forty-first International Conference on Machine Learning}, 2024.

\bibitem[Gao et~al.(2024)Gao, Holynski, Henzler, Brussee, Martin-Brualla, Srinivasan, Barron, and Poole]{gao2024cat3d}
Ruiqi Gao, Aleksander Holynski, Philipp Henzler, Arthur Brussee, Ricardo Martin-Brualla, Pratul Srinivasan, Jonathan~T Barron, and Ben Poole.
\newblock Cat3d: Create anything in 3d with multi-view diffusion models.
\newblock \emph{arXiv preprint arXiv:2405.10314}, 2024.

\bibitem[Gao et~al.(2023)Gao, Zhou, Cheng, and Yan]{gao2023mdtv2}
Shanghua Gao, Pan Zhou, Ming-Ming Cheng, and Shuicheng Yan.
\newblock Mdtv2: Masked diffusion transformer is a strong image synthesizer.
\newblock \emph{arXiv preprint arXiv:2303.14389}, 2023.

\bibitem[Geng et~al.(2024)Geng, Herrmann, Hur, Cole, Zhang, Pfaff, Lopez-Guevara, Doersch, Aytar, Rubinstein, et~al.]{geng2024motion}
Daniel Geng, Charles Herrmann, Junhwa Hur, Forrester Cole, Serena Zhang, Tobias Pfaff, Tatiana Lopez-Guevara, Carl Doersch, Yusuf Aytar, Michael Rubinstein, et~al.
\newblock Motion prompting: Controlling video generation with motion trajectories.
\newblock \emph{arXiv preprint arXiv:2412.02700}, 2024.

\bibitem[GmbH(2024)]{pexels}
Canva~Germany GmbH.
\newblock pexels, 2024.
\newblock URL https://www.pexels.com/videos/.

\bibitem[Greff et~al.(2022)Greff, Belletti, Beyer, Doersch, Du, Duckworth, Fleet, Gnanapragasam, Golemo, Herrmann, et~al.]{greff2022kubric}
Klaus Greff, Francois Belletti, Lucas Beyer, Carl Doersch, Yilun Du, Daniel Duckworth, David~J Fleet, Dan Gnanapragasam, Florian Golemo, Charles Herrmann, et~al.
\newblock Kubric: A scalable dataset generator.
\newblock In \emph{Proceedings of the IEEE/CVF conference on computer vision and pattern recognition}, pages 3749--3761, 2022.

\bibitem[Gu et~al.(2025)Gu, Yan, Lu, Li, Dou, Si, Dong, Liu, Lin, Liu, et~al.]{gu2025diffusion}
Zekai Gu, Rui Yan, Jiahao Lu, Peng Li, Zhiyang Dou, Chenyang Si, Zhen Dong, Qifeng Liu, Cheng Lin, Ziwei Liu, et~al.
\newblock Diffusion as shader: 3d-aware video diffusion for versatile video generation control.
\newblock \emph{arXiv preprint arXiv:2501.03847}, 2025.

\bibitem[He et~al.(2024)He, Xu, Guo, Wetzstein, Dai, Li, and Yang]{he2024cameractrl}
Hao He, Yinghao Xu, Yuwei Guo, Gordon Wetzstein, Bo Dai, Hongsheng Li, and Ceyuan Yang.
\newblock Cameractrl: Enabling camera control for text-to-video generation.
\newblock \emph{arXiv preprint arXiv:2404.02101}, 2024.

\bibitem[Heusel et~al.(2017)Heusel, Ramsauer, Unterthiner, Nessler, and Hochreiter]{heusel2017gans}
Martin Heusel, Hubert Ramsauer, Thomas Unterthiner, Bernhard Nessler, and Sepp Hochreiter.
\newblock Gans trained by a two time-scale update rule converge to a local nash equilibrium.
\newblock \emph{Advances in neural information processing systems}, 30, 2017.

\bibitem[Ho and Salimans(2022)]{ho2022classifier}
Jonathan Ho and Tim Salimans.
\newblock Classifier-free diffusion guidance.
\newblock \emph{arXiv preprint arXiv:2207.12598}, 2022.

\bibitem[Ho et~al.(2020)Ho, Jain, and Abbeel]{ho2020denoising}
Jonathan Ho, Ajay Jain, and Pieter Abbeel.
\newblock Denoising diffusion probabilistic models.
\newblock \emph{Advances in neural information processing systems}, 33:\penalty0 6840--6851, 2020.

\bibitem[Hou et~al.(2024)Hou, Wei, Zeng, and Chen]{hou2024training}
Chen Hou, Guoqiang Wei, Yan Zeng, and Zhibo Chen.
\newblock Training-free camera control for video generation.
\newblock \emph{arXiv preprint arXiv:2406.10126}, 2024.

\bibitem[Hu et~al.(2022)Hu, Shen, Wallis, Allen-Zhu, Li, Wang, Wang, Chen, et~al.]{hu2022lora}
Edward~J Hu, Yelong Shen, Phillip Wallis, Zeyuan Allen-Zhu, Yuanzhi Li, Shean Wang, Lu Wang, Weizhu Chen, et~al.
\newblock Lora: Low-rank adaptation of large language models.
\newblock \emph{ICLR}, 1\penalty0 (2):\penalty0 3, 2022.

\bibitem[Huang et~al.(2024{\natexlab{a}})Huang, Ghatare, Liu, Hu, Zhang, Sastry, Gururani, Oore, and Yue]{huang2024symbolic}
Yujia Huang, Adishree Ghatare, Yuanzhe Liu, Ziniu Hu, Qinsheng Zhang, Chandramouli~S Sastry, Siddharth Gururani, Sageev Oore, and Yisong Yue.
\newblock Symbolic music generation with non-differentiable rule guided diffusion.
\newblock \emph{arXiv preprint arXiv:2402.14285}, 2024{\natexlab{a}}.

\bibitem[Huang et~al.(2024{\natexlab{b}})Huang, Zhang, Xu, He, Yu, Dong, Ma, Chanpaisit, Si, Jiang, et~al.]{huang2024vbench++}
Ziqi Huang, Fan Zhang, Xiaojie Xu, Yinan He, Jiashuo Yu, Ziyue Dong, Qianli Ma, Nattapol Chanpaisit, Chenyang Si, Yuming Jiang, et~al.
\newblock Vbench++: Comprehensive and versatile benchmark suite for video generative models.
\newblock \emph{arXiv preprint arXiv:2411.13503}, 2024{\natexlab{b}}.

\bibitem[Jeong et~al.(2024{\natexlab{a}})Jeong, Chang, Park, and Ye]{jeong2024dreammotion}
Hyeonho Jeong, Jinho Chang, Geon~Yeong Park, and Jong~Chul Ye.
\newblock Dreammotion: Space-time self-similar score distillation for zero-shot video editing.
\newblock \emph{arXiv preprint arXiv:2403.12002}, 2024{\natexlab{a}}.

\bibitem[Jeong et~al.(2024{\natexlab{b}})Jeong, Huang, Ye, Mitra, and Ceylan]{jeong2024track4gen}
Hyeonho Jeong, Chun-Hao~Paul Huang, Jong~Chul Ye, Niloy Mitra, and Duygu Ceylan.
\newblock Track4gen: Teaching video diffusion models to track points improves video generation.
\newblock \emph{arXiv preprint arXiv:2412.06016}, 2024{\natexlab{b}}.

\bibitem[Jeong et~al.(2024{\natexlab{c}})Jeong, Park, and Ye]{jeong2024vmc}
Hyeonho Jeong, Geon~Yeong Park, and Jong~Chul Ye.
\newblock Vmc: Video motion customization using temporal attention adaption for text-to-video diffusion models.
\newblock In \emph{Proceedings of the IEEE/CVF Conference on Computer Vision and Pattern Recognition}, pages 9212--9221, 2024{\natexlab{c}}.

\bibitem[Jiang et~al.(2024)Jiang, Wu, Yang, Si, Lin, Qiao, Loy, and Liu]{jiang2024videobooth}
Yuming Jiang, Tianxing Wu, Shuai Yang, Chenyang Si, Dahua Lin, Yu Qiao, Chen~Change Loy, and Ziwei Liu.
\newblock Videobooth: Diffusion-based video generation with image prompts.
\newblock In \emph{Proceedings of the IEEE/CVF Conference on Computer Vision and Pattern Recognition}, pages 6689--6700, 2024.

\bibitem[Kim et~al.(2024)Kim, Kim, and Ye]{kim2024free}
Jaemin Kim, Bryan~S Kim, and Jong~Chul Ye.
\newblock Free 2 guide: Gradient-free path integral control for enhancing text-to-video generation with large vision-language models.
\newblock \emph{arXiv preprint arXiv:2411.17041}, 2024.

\bibitem[Kingma(2013)]{kingma2013auto}
Diederik~P Kingma.
\newblock Auto-encoding variational bayes.
\newblock \emph{arXiv preprint arXiv:1312.6114}, 2013.

\bibitem[Kong et~al.(2024)Kong, Tian, Zhang, Min, Dai, Zhou, Xiong, Li, Wu, Zhang, et~al.]{kong2024hunyuanvideo}
Weijie Kong, Qi Tian, Zijian Zhang, Rox Min, Zuozhuo Dai, Jin Zhou, Jiangfeng Xiong, Xin Li, Bo Wu, Jianwei Zhang, et~al.
\newblock Hunyuanvideo: A systematic framework for large video generative models.
\newblock \emph{arXiv preprint arXiv:2412.03603}, 2024.

\bibitem[Kuang et~al.(2024)Kuang, Cai, He, Xu, Li, Guibas, and Wetzstein]{kuang2024collaborative}
Zhengfei Kuang, Shengqu Cai, Hao He, Yinghao Xu, Hongsheng Li, Leonidas Guibas, and Gordon Wetzstein.
\newblock Collaborative video diffusion: Consistent multi-video generation with camera control.
\newblock \emph{arXiv preprint arXiv:2405.17414}, 2024.

\bibitem[Kuang et~al.(2025)Kuang, Cai, He, Xu, Li, Guibas, and Wetzstein]{kuang2025collaborative}
Zhengfei Kuang, Shengqu Cai, Hao He, Yinghao Xu, Hongsheng Li, Leonidas~J Guibas, and Gordon Wetzstein.
\newblock Collaborative video diffusion: Consistent multi-video generation with camera control.
\newblock \emph{Advances in Neural Information Processing Systems}, 37:\penalty0 16240--16271, 2025.

\bibitem[Kulikov et~al.(2024)Kulikov, Kleiner, Huberman-Spiegelglas, and Michaeli]{kulikov2024flowedit}
Vladimir Kulikov, Matan Kleiner, Inbar Huberman-Spiegelglas, and Tomer Michaeli.
\newblock Flowedit: Inversion-free text-based editing using pre-trained flow models.
\newblock \emph{arXiv preprint arXiv:2412.08629}, 2024.

\bibitem[Kwon and Ye(2024)]{kwon2024tweediemix}
Gihyun Kwon and Jong~Chul Ye.
\newblock Tweediemix: Improving multi-concept fusion for diffusion-based image/video generation.
\newblock \emph{arXiv preprint arXiv:2410.05591}, 2024.

\bibitem[Labs(2024)]{flux}
Black~Forest Labs.
\newblock Flux, 2024.
\newblock URL https://blackforestlabs.ai/.

\bibitem[Li et~al.(2024)Li, Zheng, Zhu, Mai, Zhang, Wonka, and Ghanem]{li2024vivid}
Bing Li, Cheng Zheng, Wenxuan Zhu, Jinjie Mai, Biao Zhang, Peter Wonka, and Bernard Ghanem.
\newblock Vivid-zoo: Multi-view video generation with diffusion model.
\newblock \emph{arXiv preprint arXiv:2406.08659}, 2024.

\bibitem[Li et~al.(2025)Li, Zheng, Jiang, Wu, Lu, Lin, Li, et~al.]{li2025realcam}
Teng Li, Guangcong Zheng, Rui Jiang, Tao Wu, Yehao Lu, Yining Lin, Xi Li, et~al.
\newblock Realcam-i2v: Real-world image-to-video generation with interactive complex camera control.
\newblock \emph{arXiv preprint arXiv:2502.10059}, 2025.

\bibitem[Liang et~al.(2024)Liang, Yin, Xu, Liang, Wang, Plataniotis, Zhao, and Wei]{liang2024diffusion4d}
Hanwen Liang, Yuyang Yin, Dejia Xu, Hanxue Liang, Zhangyang Wang, Konstantinos~N Plataniotis, Yao Zhao, and Yunchao Wei.
\newblock Diffusion4d: Fast spatial-temporal consistent 4d generation via video diffusion models.
\newblock \emph{arXiv preprint arXiv:2405.16645}, 2024.

\bibitem[Ling et~al.(2024)Ling, Sheng, Tu, Zhao, Xin, Wan, Yu, Guo, Yu, Lu, et~al.]{ling2024dl3dv}
Lu Ling, Yichen Sheng, Zhi Tu, Wentian Zhao, Cheng Xin, Kun Wan, Lantao Yu, Qianyu Guo, Zixun Yu, Yawen Lu, et~al.
\newblock Dl3dv-10k: A large-scale scene dataset for deep learning-based 3d vision.
\newblock In \emph{Proceedings of the IEEE/CVF Conference on Computer Vision and Pattern Recognition}, pages 22160--22169, 2024.

\bibitem[Liu(2024)]{liu_flow_together}
Qiang Liu.
\newblock \emph{Let Us Flow Together}.
\newblock 2024.
\newblock Accessed: 2025-02-26.

\bibitem[Loshchilov(2017)]{loshchilov2017decoupled}
I Loshchilov.
\newblock Decoupled weight decay regularization.
\newblock \emph{arXiv preprint arXiv:1711.05101}, 2017.

\bibitem[Park et~al.(2024)Park, Jeong, Lee, and Ye]{park2024spectral}
Geon~Yeong Park, Hyeonho Jeong, Sang~Wan Lee, and Jong~Chul Ye.
\newblock Spectral motion alignment for video motion transfer using diffusion models.
\newblock \emph{arXiv preprint arXiv:2403.15249}, 2024.

\bibitem[Peebles and Xie(2023)]{peebles2023scalable}
William Peebles and Saining Xie.
\newblock Scalable diffusion models with transformers.
\newblock In \emph{Proceedings of the IEEE/CVF International Conference on Computer Vision}, pages 4195--4205, 2023.

\bibitem[Polyak et~al.(2024)Polyak, Zohar, Brown, Tjandra, Sinha, Lee, Vyas, Shi, Ma, Chuang, Yan, Choudhary, Wang, Sethi, Pang, Ma, Misra, Hou, Wang, Jagadeesh, Li, Zhang, Singh, Williamson, Le, Yu, Singh, Zhang, Vajda, Duval, Girdhar, Sumbaly, Rambhatla, Tsai, Azadi, Datta, Chen, Bell, Ramaswamy, Sheynin, Bhattacharya, Motwani, Xu, Li, Hou, Hsu, Yin, Dai, Taigman, Luo, Liu, Wu, Zhao, Kirstain, He, He, Pumarola, Thabet, Sanakoyeu, Mallya, Guo, Araya, Kerr, Wood, Liu, Peng, Vengertsev, Schonfeld, Blanchard, Juefei-Xu, Nord, Liang, Hoffman, Kohler, Fire, Sivakumar, Chen, Yu, Gao, Georgopoulos, Moritz, Sampson, Li, Parmeggiani, Fine, Fowler, Petrovic, and Du]{polyak2024moviegencastmedia}
Adam Polyak, Amit Zohar, Andrew Brown, Andros Tjandra, Animesh Sinha, Ann Lee, Apoorv Vyas, Bowen Shi, Chih-Yao Ma, Ching-Yao Chuang, David Yan, Dhruv Choudhary, Dingkang Wang, Geet Sethi, Guan Pang, Haoyu Ma, Ishan Misra, Ji Hou, Jialiang Wang, Kiran Jagadeesh, Kunpeng Li, Luxin Zhang, Mannat Singh, Mary Williamson, Matt Le, Matthew Yu, Mitesh~Kumar Singh, Peizhao Zhang, Peter Vajda, Quentin Duval, Rohit Girdhar, Roshan Sumbaly, Sai~Saketh Rambhatla, Sam Tsai, Samaneh Azadi, Samyak Datta, Sanyuan Chen, Sean Bell, Sharadh Ramaswamy, Shelly Sheynin, Siddharth Bhattacharya, Simran Motwani, Tao Xu, Tianhe Li, Tingbo Hou, Wei-Ning Hsu, Xi Yin, Xiaoliang Dai, Yaniv Taigman, Yaqiao Luo, Yen-Cheng Liu, Yi-Chiao Wu, Yue Zhao, Yuval Kirstain, Zecheng He, Zijian He, Albert Pumarola, Ali Thabet, Artsiom Sanakoyeu, Arun Mallya, Baishan Guo, Boris Araya, Breena Kerr, Carleigh Wood, Ce Liu, Cen Peng, Dimitry Vengertsev, Edgar Schonfeld, Elliot Blanchard, Felix Juefei-Xu, Fraylie Nord, Jeff Liang, John Hoffman, Jonas
  Kohler, Kaolin Fire, Karthik Sivakumar, Lawrence Chen, Licheng Yu, Luya Gao, Markos Georgopoulos, Rashel Moritz, Sara~K. Sampson, Shikai Li, Simone Parmeggiani, Steve Fine, Tara Fowler, Vladan Petrovic, and Yuming Du.
\newblock Movie gen: A cast of media foundation models, 2024.

\bibitem[Pont-Tuset et~al.(2017)Pont-Tuset, Perazzi, Caelles, Arbel{\'a}ez, Sorkine-Hornung, and Van~Gool]{pont20172017}
Jordi Pont-Tuset, Federico Perazzi, Sergi Caelles, Pablo Arbel{\'a}ez, Alex Sorkine-Hornung, and Luc Van~Gool.
\newblock The 2017 davis challenge on video object segmentation.
\newblock \emph{arXiv preprint arXiv:1704.00675}, 2017.

\bibitem[Popov et~al.(2025)Popov, Raj, Krainin, Li, Freeman, and Rubinstein]{popov2025camctrl3d}
Stefan Popov, Amit Raj, Michael Krainin, Yuanzhen Li, William~T Freeman, and Michael Rubinstein.
\newblock Camctrl3d: Single-image scene exploration with precise 3d camera control.
\newblock \emph{arXiv preprint arXiv:2501.06006}, 2025.

\bibitem[Rombach et~al.(2021)Rombach, Blattmann, Lorenz, Esser, and Ommer]{Rombach2021HighResolutionIS}
Robin Rombach, A. Blattmann, Dominik Lorenz, Patrick Esser, and Bj{\"o}rn Ommer.
\newblock High-resolution image synthesis with latent diffusion models.
\newblock In \emph{CVPR}, 2021.

\bibitem[Ryu(2023)]{ryu2023low}
Simo Ryu.
\newblock Low-rank adaptation for fast text-to-image diffusion fine-tuning.
\newblock \emph{Low-rank adaptation for fast text-to-image diffusion fine-tuning}, 3, 2023.

\bibitem[Sanders(2016)]{sanders2016introduction}
Andrew Sanders.
\newblock \emph{An introduction to Unreal engine 4}.
\newblock AK Peters/CRC Press, 2016.

\bibitem[Shao et~al.(2024)Shao, Pang, Zheng, Sun, and Liu]{shao2024360}
Ruizhi Shao, Youxin Pang, Zerong Zheng, Jingxiang Sun, and Yebin Liu.
\newblock 360-degree human video generation with 4d diffusion transformer.
\newblock \emph{ACM Transactions on Graphics (TOG)}, 43\penalty0 (6):\penalty0 1--13, 2024.

\bibitem[Singh and Fischer(2024)]{singh2024stochastic}
Saurabh Singh and Ian Fischer.
\newblock Stochastic sampling from deterministic flow models.
\newblock \emph{arXiv preprint arXiv:2410.02217}, 2024.

\bibitem[Sohl-Dickstein et~al.(2015)Sohl-Dickstein, Weiss, Maheswaranathan, and Ganguli]{sohl2015deep}
Jascha Sohl-Dickstein, Eric Weiss, Niru Maheswaranathan, and Surya Ganguli.
\newblock Deep unsupervised learning using nonequilibrium thermodynamics.
\newblock In \emph{International conference on machine learning}, pages 2256--2265. PMLR, 2015.

\bibitem[Song et~al.(2020)Song, Sohl-Dickstein, Kingma, Kumar, Ermon, and Poole]{song2020score}
Yang Song, Jascha Sohl-Dickstein, Diederik~P Kingma, Abhishek Kumar, Stefano Ermon, and Ben Poole.
\newblock Score-based generative modeling through stochastic differential equations.
\newblock \emph{arXiv preprint arXiv:2011.13456}, 2020.

\bibitem[Sun et~al.(2024)Sun, Chen, Liu, Chen, Duan, Zhang, and Wang]{sun2024dimensionx}
Wenqiang Sun, Shuo Chen, Fangfu Liu, Zilong Chen, Yueqi Duan, Jun Zhang, and Yikai Wang.
\newblock Dimensionx: Create any 3d and 4d scenes from a single image with controllable video diffusion.
\newblock \emph{arXiv preprint arXiv:2411.04928}, 2024.

\bibitem[Unterthiner et~al.(2018)Unterthiner, Van~Steenkiste, Kurach, Marinier, Michalski, and Gelly]{unterthiner2018towards}
Thomas Unterthiner, Sjoerd Van~Steenkiste, Karol Kurach, Raphael Marinier, Marcin Michalski, and Sylvain Gelly.
\newblock Towards accurate generative models of video: A new metric \& challenges.
\newblock \emph{arXiv preprint arXiv:1812.01717}, 2018.

\bibitem[Van~Hoorick et~al.(2024)Van~Hoorick, Wu, Ozguroglu, Sargent, Liu, Tokmakov, Dave, Zheng, and Vondrick]{van2024generative}
Basile Van~Hoorick, Rundi Wu, Ege Ozguroglu, Kyle Sargent, Ruoshi Liu, Pavel Tokmakov, Achal Dave, Changxi Zheng, and Carl Vondrick.
\newblock Generative camera dolly: Extreme monocular dynamic novel view synthesis.
\newblock In \emph{European Conference on Computer Vision}, pages 313--331. Springer, 2024.

\bibitem[Wallace et~al.(2023)Wallace, Gokul, Ermon, and Naik]{wallace2023end}
Bram Wallace, Akash Gokul, Stefano Ermon, and Nikhil Naik.
\newblock End-to-end diffusion latent optimization improves classifier guidance.
\newblock In \emph{Proceedings of the IEEE/CVF International Conference on Computer Vision}, pages 7280--7290, 2023.

\bibitem[Wang et~al.(2024{\natexlab{a}})Wang, Zhuang, Ngo, Menapace, Siarohin, Vasilkovsky, Skorokhodov, Tulyakov, Wonka, and Lee]{wang20244real}
Chaoyang Wang, Peiye Zhuang, Tuan~Duc Ngo, Willi Menapace, Aliaksandr Siarohin, Michael Vasilkovsky, Ivan Skorokhodov, Sergey Tulyakov, Peter Wonka, and Hsin-Ying Lee.
\newblock 4real-video: Learning generalizable photo-realistic 4d video diffusion.
\newblock \emph{arXiv preprint arXiv:2412.04462}, 2024{\natexlab{a}}.

\bibitem[Wang et~al.(2024{\natexlab{b}})Wang, Leroy, Cabon, Chidlovskii, and Revaud]{wang2024dust3r}
Shuzhe Wang, Vincent Leroy, Yohann Cabon, Boris Chidlovskii, and Jerome Revaud.
\newblock Dust3r: Geometric 3d vision made easy.
\newblock In \emph{Proceedings of the IEEE/CVF Conference on Computer Vision and Pattern Recognition}, pages 20697--20709, 2024{\natexlab{b}}.

\bibitem[Wang et~al.(2024{\natexlab{c}})Wang, Chen, Ma, Zhou, Huang, Wang, Yang, He, Yu, Yang, et~al.]{wang2024lavie}
Yaohui Wang, Xinyuan Chen, Xin Ma, Shangchen Zhou, Ziqi Huang, Yi Wang, Ceyuan Yang, Yinan He, Jiashuo Yu, Peiqing Yang, et~al.
\newblock Lavie: High-quality video generation with cascaded latent diffusion models.
\newblock \emph{International Journal of Computer Vision}, pages 1--20, 2024{\natexlab{c}}.

\bibitem[Wang et~al.(2024{\natexlab{d}})Wang, Yuan, Wang, Li, Chen, Xia, Luo, and Shan]{wang2024motionctrl}
Zhouxia Wang, Ziyang Yuan, Xintao Wang, Yaowei Li, Tianshui Chen, Menghan Xia, Ping Luo, and Ying Shan.
\newblock Motionctrl: A unified and flexible motion controller for video generation.
\newblock In \emph{ACM SIGGRAPH 2024 Conference Papers}, pages 1--11, 2024{\natexlab{d}}.

\bibitem[Watson et~al.(2024)Watson, Saxena, Li, Tagliasacchi, and Fleet]{watson2024controlling}
Daniel Watson, Saurabh Saxena, Lala Li, Andrea Tagliasacchi, and David~J Fleet.
\newblock Controlling space and time with diffusion models.
\newblock \emph{arXiv preprint arXiv:2407.07860}, 2024.

\bibitem[Wei et~al.(2024)Wei, Zhang, Yuan, Wang, Qiu, Zhao, Feng, Liu, Huang, Ye, et~al.]{wei2024dreamvideo2}
Yujie Wei, Shiwei Zhang, Hangjie Yuan, Xiang Wang, Haonan Qiu, Rui Zhao, Yutong Feng, Feng Liu, Zhizhong Huang, Jiaxin Ye, et~al.
\newblock Dreamvideo-2: Zero-shot subject-driven video customization with precise motion control.
\newblock \emph{arXiv preprint arXiv:2410.13830}, 2024.

\bibitem[Wu et~al.(2024)Wu, Gao, Poole, Trevithick, Zheng, Barron, and Holynski]{wu2024cat4d}
Rundi Wu, Ruiqi Gao, Ben Poole, Alex Trevithick, Changxi Zheng, Jonathan~T Barron, and Aleksander Holynski.
\newblock Cat4d: Create anything in 4d with multi-view video diffusion models.
\newblock \emph{arXiv preprint arXiv:2411.18613}, 2024.

\bibitem[Xiao et~al.(2024)Xiao, Ouyang, Zhou, Yang, Yang, Si, and Pan]{xiao2024trajectory}
Zeqi Xiao, Wenqi Ouyang, Yifan Zhou, Shuai Yang, Lei Yang, Jianlou Si, and Xingang Pan.
\newblock Trajectory attention for fine-grained video motion control.
\newblock \emph{arXiv preprint arXiv:2411.19324}, 2024.

\bibitem[Xie et~al.(2024)Xie, Yao, Voleti, Jiang, and Jampani]{xie2024sv4d}
Yiming Xie, Chun-Han Yao, Vikram Voleti, Huaizu Jiang, and Varun Jampani.
\newblock Sv4d: Dynamic 3d content generation with multi-frame and multi-view consistency.
\newblock \emph{arXiv preprint arXiv:2407.17470}, 2024.

\bibitem[Xu et~al.(2024{\natexlab{a}})Xu, Jiang, Huang, Song, Gernoth, Cao, Wang, and Tang]{xu2024cavia}
Dejia Xu, Yifan Jiang, Chen Huang, Liangchen Song, Thorsten Gernoth, Liangliang Cao, Zhangyang Wang, and Hao Tang.
\newblock Cavia: Camera-controllable multi-view video diffusion with view-integrated attention.
\newblock \emph{arXiv preprint arXiv:2410.10774}, 2024{\natexlab{a}}.

\bibitem[Xu et~al.(2024{\natexlab{b}})Xu, Nie, Liu, Liu, Kautz, Wang, and Vahdat]{xu2024camco}
Dejia Xu, Weili Nie, Chao Liu, Sifei Liu, Jan Kautz, Zhangyang Wang, and Arash Vahdat.
\newblock Camco: Camera-controllable 3d-consistent image-to-video generation.
\newblock \emph{arXiv preprint arXiv:2406.02509}, 2024{\natexlab{b}}.

\bibitem[Yang et~al.(2024{\natexlab{a}})Yang, Kang, Huang, Zhao, Xu, Feng, and Zhao]{yang2024depth}
Lihe Yang, Bingyi Kang, Zilong Huang, Zhen Zhao, Xiaogang Xu, Jiashi Feng, and Hengshuang Zhao.
\newblock Depth anything v2.
\newblock \emph{arXiv preprint arXiv:2406.09414}, 2024{\natexlab{a}}.

\bibitem[Yang et~al.(2024{\natexlab{b}})Yang, Hou, Huang, Ma, Wan, Zhang, Chen, and Liao]{yang2024direct}
Shiyuan Yang, Liang Hou, Haibin Huang, Chongyang Ma, Pengfei Wan, Di Zhang, Xiaodong Chen, and Jing Liao.
\newblock Direct-a-video: Customized video generation with user-directed camera movement and object motion.
\newblock In \emph{ACM SIGGRAPH 2024 Conference Papers}, pages 1--12, 2024{\natexlab{b}}.

\bibitem[Yang et~al.(2024{\natexlab{c}})Yang, Teng, Zheng, Ding, Huang, Xu, Yang, Hong, Zhang, Feng, et~al.]{yang2024cogvideox}
Zhuoyi Yang, Jiayan Teng, Wendi Zheng, Ming Ding, Shiyu Huang, Jiazheng Xu, Yuanming Yang, Wenyi Hong, Xiaohan Zhang, Guanyu Feng, et~al.
\newblock Cogvideox: Text-to-video diffusion models with an expert transformer.
\newblock \emph{arXiv preprint arXiv:2408.06072}, 2024{\natexlab{c}}.

\bibitem[Yatim et~al.(2024)Yatim, Fridman, Bar-Tal, Kasten, and Dekel]{yatim2024space}
Danah Yatim, Rafail Fridman, Omer Bar-Tal, Yoni Kasten, and Tali Dekel.
\newblock Space-time diffusion features for zero-shot text-driven motion transfer.
\newblock In \emph{Proceedings of the IEEE/CVF Conference on Computer Vision and Pattern Recognition}, pages 8466--8476, 2024.

\bibitem[Yeh et~al.(2024)Yeh, Lee, and Chen]{yeh2024training}
Po-Hung Yeh, Kuang-Huei Lee, and Jun-Cheng Chen.
\newblock Training-free diffusion model alignment with sampling demons.
\newblock \emph{arXiv preprint arXiv:2410.05760}, 2024.

\bibitem[You et~al.(2024)You, Zhu, Liu, and Hou]{you2024nvs}
Meng You, Zhiyu Zhu, Hui Liu, and Junhui Hou.
\newblock Nvs-solver: Video diffusion model as zero-shot novel view synthesizer.
\newblock \emph{arXiv preprint arXiv:2405.15364}, 2024.

\bibitem[Yu et~al.(2023)Yu, Forghani, Derpanis, and Brubaker]{yu2023long}
Jason~J Yu, Fereshteh Forghani, Konstantinos~G Derpanis, and Marcus~A Brubaker.
\newblock Long-term photometric consistent novel view synthesis with diffusion models.
\newblock In \emph{Proceedings of the IEEE/CVF International Conference on Computer Vision}, pages 7094--7104, 2023.

\bibitem[Yu et~al.(2024)Yu, Xing, Yuan, Hu, Li, Huang, Gao, Wong, Shan, and Tian]{yu2024viewcrafter}
Wangbo Yu, Jinbo Xing, Li Yuan, Wenbo Hu, Xiaoyu Li, Zhipeng Huang, Xiangjun Gao, Tien-Tsin Wong, Ying Shan, and Yonghong Tian.
\newblock Viewcrafter: Taming video diffusion models for high-fidelity novel view synthesis.
\newblock \emph{arXiv preprint arXiv:2409.02048}, 2024.

\bibitem[Zhang et~al.(2024{\natexlab{a}})Zhang, Paiss, Zada, Karnad, Jacobs, Pritch, Mosseri, Shou, Wadhwa, and Ruiz]{zhang2024recapture}
David~Junhao Zhang, Roni Paiss, Shiran Zada, Nikhil Karnad, David~E Jacobs, Yael Pritch, Inbar Mosseri, Mike~Zheng Shou, Neal Wadhwa, and Nataniel Ruiz.
\newblock Recapture: Generative video camera controls for user-provided videos using masked video fine-tuning.
\newblock \emph{arXiv preprint arXiv:2411.05003}, 2024{\natexlab{a}}.

\bibitem[Zhang et~al.(2024{\natexlab{b}})Zhang, Wu, Liu, Zhao, Ran, Gu, Gao, and Shou]{zhang2024show}
David~Junhao Zhang, Jay~Zhangjie Wu, Jia-Wei Liu, Rui Zhao, Lingmin Ran, Yuchao Gu, Difei Gao, and Mike~Zheng Shou.
\newblock Show-1: Marrying pixel and latent diffusion models for text-to-video generation.
\newblock \emph{International Journal of Computer Vision}, pages 1--15, 2024{\natexlab{b}}.

\bibitem[Zhang et~al.(2023)Zhang, Rao, and Agrawala]{zhang2023adding}
Lvmin Zhang, Anyi Rao, and Maneesh Agrawala.
\newblock Adding conditional control to text-to-image diffusion models.
\newblock In \emph{Proceedings of the IEEE/CVF International Conference on Computer Vision}, pages 3836--3847, 2023.

\bibitem[Zhao et~al.(2024)Zhao, Gu, Wu, Zhang, Liu, Wu, Keppo, and Shou]{zhao2024motiondirector}
Rui Zhao, Yuchao Gu, Jay~Zhangjie Wu, David~Junhao Zhang, Jia-Wei Liu, Weijia Wu, Jussi Keppo, and Mike~Zheng Shou.
\newblock Motiondirector: Motion customization of text-to-video diffusion models.
\newblock In \emph{European Conference on Computer Vision}, pages 273--290. Springer, 2024.

\bibitem[Zheng et~al.(2023)Zheng, Nie, Vahdat, and Anandkumar]{zheng2023fast}
Hongkai Zheng, Weili Nie, Arash Vahdat, and Anima Anandkumar.
\newblock Fast training of diffusion models with masked transformers.
\newblock \emph{arXiv preprint arXiv:2306.09305}, 2023.

\bibitem[Zhou et~al.(2018)Zhou, Tucker, Flynn, Fyffe, and Snavely]{zhou2018stereo}
Tinghui Zhou, Richard Tucker, John Flynn, Graham Fyffe, and Noah Snavely.
\newblock Stereo magnification: Learning view synthesis using multiplane images.
\newblock \emph{arXiv preprint arXiv:1805.09817}, 2018.

\bibitem[Zuo et~al.(2024)Zuo, Gu, Qiu, Dong, Zhao, Yuan, Peng, Zhu, Dong, Bo, et~al.]{zuo2024videomv}
Qi Zuo, Xiaodong Gu, Lingteng Qiu, Yuan Dong, Zhengyi Zhao, Weihao Yuan, Rui Peng, Siyu Zhu, Zilong Dong, Liefeng Bo, et~al.
\newblock Videomv: Consistent multi-view generation based on large video generative model.
\newblock \emph{arXiv preprint arXiv:2403.12010}, 2024.

\end{thebibliography}
}

\clearpage
\setcounter{page}{1}
\maketitlesupplementary

\appendix

This supplementary material is structured as follows:
In Sec. \ref{sup: exp-details}, we provide additional experimental details.
Sec. \ref{sup: exp-results} presents additional experiment results.
Following this, we discuss the limitations and failure cases of Reangle-A-Video in Sec. \ref{sup: limitations}.

\section{Additional Experimental Details}
\label{sup: exp-details}

\subsection{Camera Visualizations}
\label{sup: sec-camera-types}
\begin{figure}[!h]
    \centering
    \includegraphics[width=\columnwidth]
    {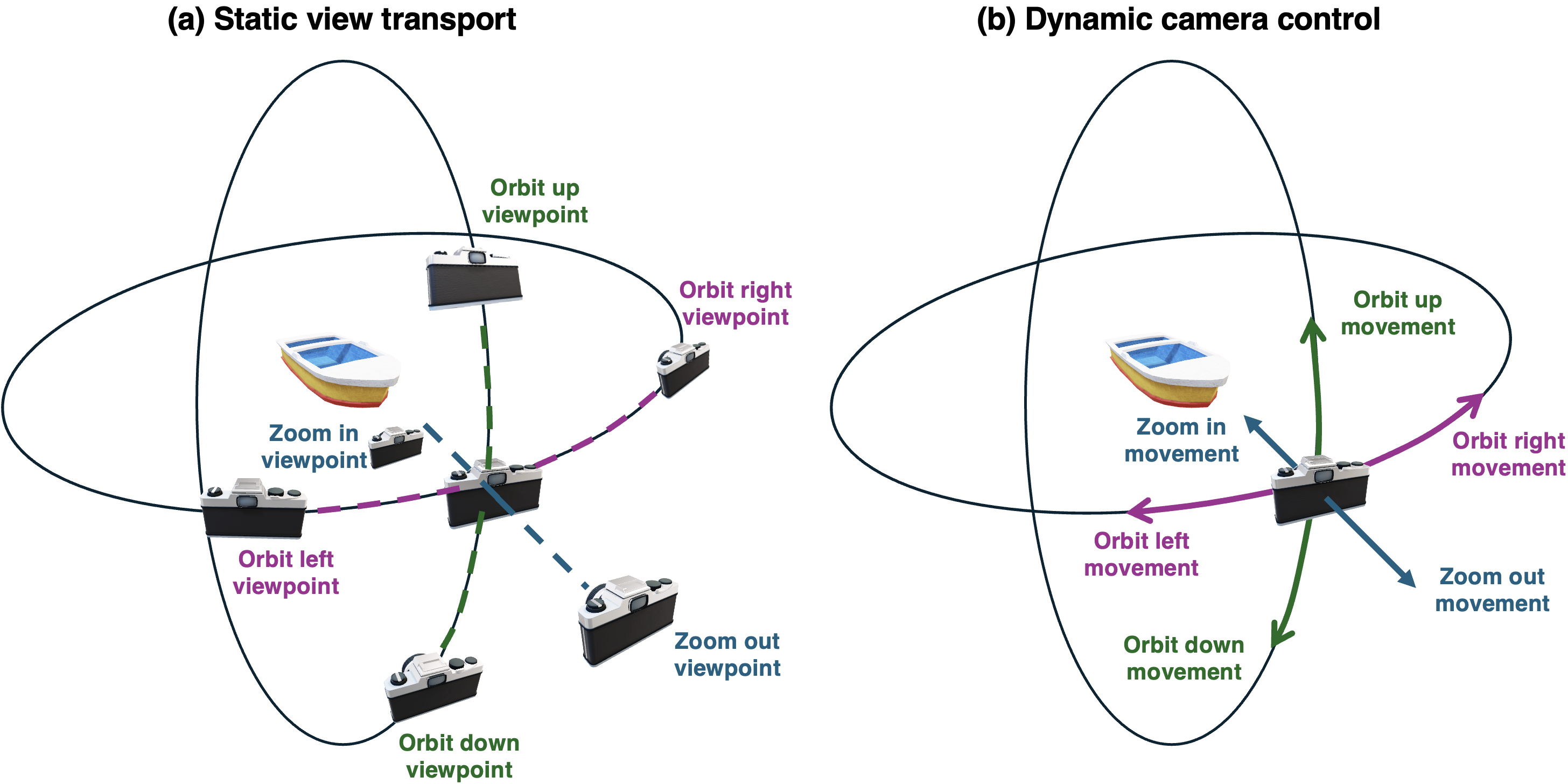}
    \caption{
    \textbf{Visualizations of the used camera types.}
    }
    \label{appendix: camera-vis}
\end{figure}

We demonstrate six degrees of freedom in both (a) Static view transport and (b) Dynamic camera control.
Fig. \ref{appendix: camera-vis} visualizes the transported viewpoints and camera movements used in our work: 
\textit{orbit left, orbit right, orbit up, orbit down, dolly zoom in, and dolly zoom out.}

\subsection{Warped Video Dataset Composition}
\label{sup: training dataset composition}
For (a) static view transport setup, we set $M$ as 12, generating warped videos from 12 different viewpoints:
two random-angle orbit left, two random-angle orbit right, two random-angle orbit up, two random-angle orbit down, two random-range dolly zoom in, and two random-range dolly zoom out.
Including the original input video, the training set consists of 13 videos.

For (b) dynamic camera control training, we set $M$ as 6, rendering warped videos with six different camera movements:
one random-angle orbit left, one random-angle orbit right, one random-angle orbit up, one random-angle orbit down, one random-range dolly zoom in, and one random-range dolly zoom out.
Including the original input video, this also results in a training set of 7 videos.

\subsection{Finetuning Cost}
We  originally trained and inferred at a high resolution ($49{\times}480{\times}720$) on a 40GB GPU using \textit{gradient checkpointing} to fit within memory—this significantly slows down backpropagation.
However, on an 80GB GPU without gradient checkpointing, the process is 3.3${\times}$ faster, completing in about 18 minutes.
Alternatively, on a 40GB GPU with checkpointing but with reduced temporal resolution (25 frames), it’s 2.3${\times}$ faster (26 minutes).

\subsection{3D Downsampling Visiblity Masks}
\label{sup: 3d downsampling masks}
\begin{figure}[!t]
    \centering
    \includegraphics[width=\columnwidth]
    {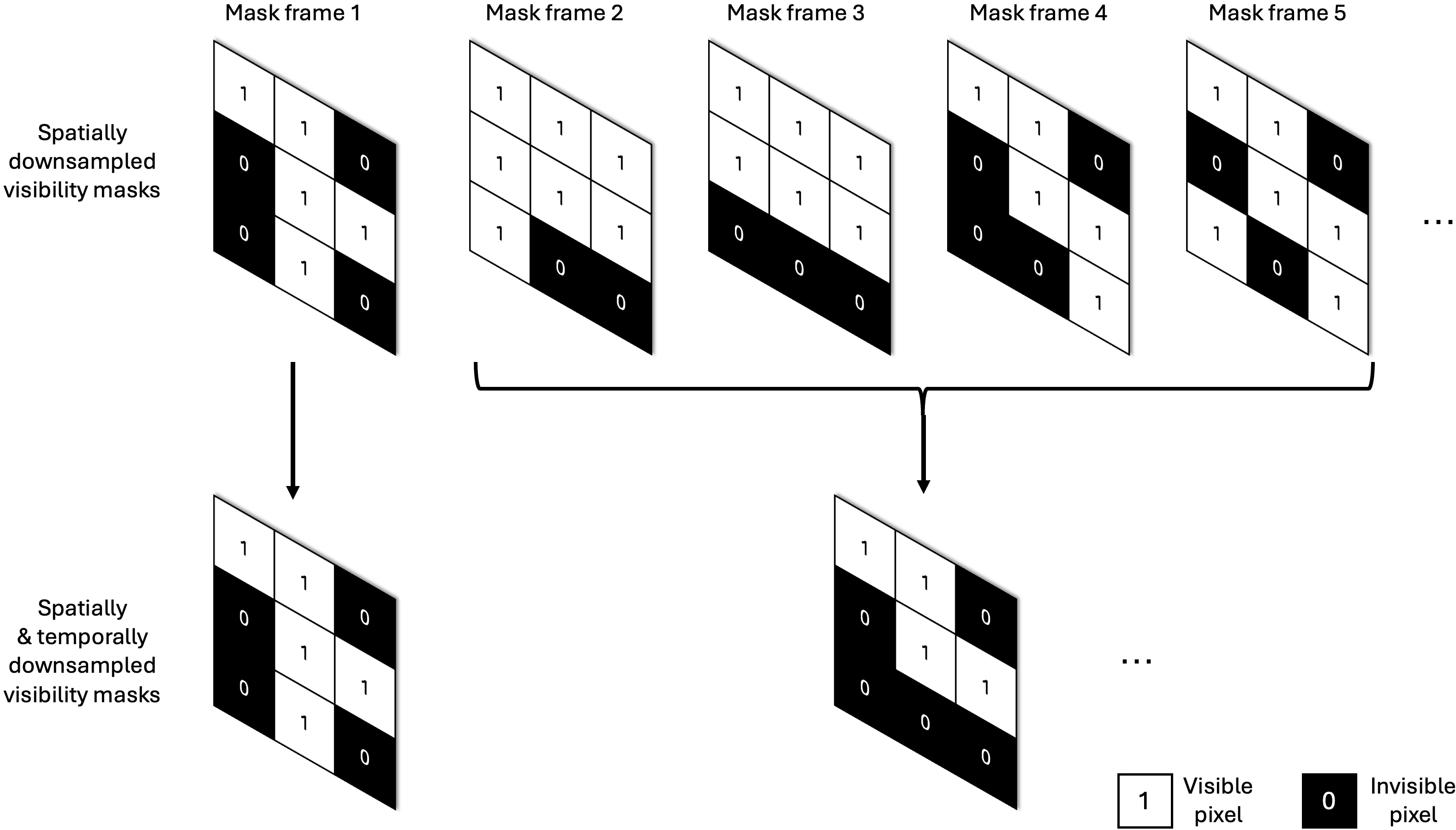}
    \caption{
    \textbf{Temporal downsampling of visibility masks.}
    Except for the first mask frame, pixel-wise (element-wise) logical AND operation is done for every four masks.
    }
    \label{appendix: mask-downsampling}
\end{figure}

Recent video diffusion models rely on 3D VAEs that perform both spatial and temporal compression to alleviate the computational burden of modeling long video sequences. 
However, such compression complicates the direct mapping of RGB pixel-space visibility masks to their corresponding latent regions. 
In our work, we adopt the 3D VAE architecture of CogVideoX, which features a spatio-temporal compression rate given by $H \times W \times N = 8h \times 8w \times (1+4n)$, where $(H, W, N)$ denote the pixel-space resolutions, and $(h, w, n)$  represent the corresponding resolutions in the latent space.
Given a visibility mask in pixel-space, $M \in \mathbb{R}^{H \times W \times N}$, we first downsample the spatial dimensions by a factor of 8 using \textit{nearest-neighbor} interpolation. 
For temporal downsampling, we retain the first frame intact and then compress every subsequent group of four frames via an element-wise logical AND operation (see Fig. \ref{appendix: mask-downsampling} for the illustration). This procedure ensures that a latent region is marked as visible only if it is visible across all frames within each group.

\subsection{Pre-trained Model Checkpoints}
\label{sup: model ckpts}
Reangle-A-Video builds upon publicly available pre-trained image and video generative models.
Reangle-A-Video builds on publicly available pre-trained image and video generative models. Here, we specify the versions used:
\begin{itemize}
    \item Text-to-Image generation model: Flux.1-dev \footnote{\url{https://huggingface.co/black-forest-labs/FLUX.1-dev}}
    
    \item Image-to-Image inpainting model: Flux-ControlNet-Inpainting-Beta \footnote{\url{https://huggingface.co/alimama-creative/FLUX.1-dev-Controlnet-Inpainting-Beta}}

    \item Image-to-Video generation model: CogVideoX-I2V-5b \footnote{\url{https://huggingface.co/THUDM/CogVideoX-5b-I2V}}
\end{itemize}

\begin{algorithm}
\caption{Multi-View Consistent Image Inpainting (diffusion model)}\label{alg:inpaint-diff}
\label{supp: alg-diffusion}
\begin{algorithmic}
\Require Inpainted images $\{\tilde{\xb}_1, \dots, \tilde{\xb}_I\}$, the image to be inpainted $\hat{\xb}_{I+1}$, and the image inpainting diffusion model $(\phi, \mathcal{E}, \mathcal{D})$.
\State $\cb \gets \mathcal{E}(\hat{\xb}_{I+1})$
\State $\zb_{T}\sim\mathcal{N}(\textbf{0},\textbf{I})$
\For{$t=T$ \textbf{to} 1}
\For{$s=1$ \textbf{to} $S$}
\State $\epsilonb\sim\mathcal{N}(\textbf{0},\textbf{I})$
\State $\zb^s_{t-1} \gets \text{DDPM\_step}(\zb_{t},\phi, \cb, t,\epsilonb)$
\State $\smash{ \zb_{0|t-1}^s \gets \frac{1}{\sqrt{\bar{\alpha}_{t-1}}} (\zb_{t-1}^s-\sqrt{1-\bar{\alpha}_{t-1}}\Hat{\epsilonb}_\phi(\zb_{t-1}^s,\cb,t)) }$
\EndFor

\For{$s=1$ \textbf{to} $S$}
\For{$j=1$ \textbf{to} $I$}
\State $r_{s,j} \gets \text{eval\_3d\_consistency}(\mathcal{D}(\zb_{0|t-1}^s), \Tilde{\xb}_j)$
\EndFor
\State $r_s \gets\frac{1}{I}(r_{s,1}+r_{s,2}+\cdots+r_{s,I})$
\EndFor
\State $s^* \gets \text{argmax}_s \; r_s$
\State $\zb_{t-1} \gets \zb_{t-1}^{s^*}$
\EndFor
\State $\Tilde{\xb}_{I+1} \gets \mathcal{D}(\zb_0)$ \\
\Return {$\Tilde{\xb}_{I+1}$}
\end{algorithmic}
\end{algorithm}

\subsection{Multi-view Consistent Images Completion}
\label{supp: sec: inpainting details}
We present the detailed algorithm for our proposed multi-view consistent image completion. As described in the main text, our approach sequentially performs multi-view inpainting using Algorithm~\ref{alg:inpaint-diff} or \ref{alg:inpaint-flow}, based on previously completed images. We introduce two versions: a diffusion model-based method (Algorithm 1) and a flow model-based method (Algorithm 2).

To generate multiple sample versions at each denoising step for the stochastic control guidance~\cite{singh2024stochastic, kim2024free, yeh2024training}, the diffusion model-based approach employs DDPM~\cite{ho2020denoising} sampling, while the flow model-based approach solves an SDE~\cite{liu_flow_together} that shares the same marginal distribution as its corresponding ODE. Specifically, for the Wiener process $W_t$:
\begin{multline}
dZ_t = \vb_t(Z_t)dt - Z_{0|t} (1-t)dt + \sqrt{2(1-t)^2}dW_t, \\
\text{where} \quad Z_{0|t} = Z_t - t \vb_t(Z_t).
\end{multline}
The SDE is solved using the Euler–Maruyama method.

\begin{algorithm}
\caption{Multi-View Consistent Image Inpainting (flow-based model)}\label{alg:inpaint-flow}
\label{supp: alg-flow}
\begin{algorithmic}
\Require Inpainted images $\{\tilde{\xb}_1, \dots, \tilde{\xb}_I\}$, the image to be inpainted $\hat{\xb}_{I+1}$, and the image inpainting flow-based model $(\phi, \mathcal{E}, \mathcal{D})$.
\State $\cb \gets \mathcal{E}(\hat{\xb}_{I+1})$
\State $\zb_{t_1}\sim\mathcal{N}(\textbf{0},\textbf{I})$
\For{$i=0$ \textbf{to} $T-1$}
\State $\Delta t \gets t_{i+1} - t_i$
\State $\zb_{0|t_i} \gets \zb_{t_i} - t_i \hat{\vb}_\phi(\zb_{t_{i}},\cb,t_i)$
\For{$s=1$ \textbf{to} $S$}
\State $\epsilonb\sim\mathcal{N}(\textbf{0},\textbf{I})$
\State $\zb^s_{t_{i+1}} \gets \zb_{t_{i}} + \hat{\vb}_\phi(\zb_{t_{i}},\cb,t_i) \Delta t $
\State $\qquad \qquad \quad \; - \zb_{0|t_i} (1-t_i) \Delta t + \sqrt{2(1-t_i)^2\Delta t} \epsilonb$
\State $\zb_{1|t_{i+1}}^s \gets \zb_{t_{i+1}}^s + (1-t) \hat{\vb}_\phi(\zb_{t_{i+1}}^s,\cb,t_{i+1})  $
\EndFor

\For{$s=1$ \textbf{to} $S$}
\For{$j=1$ \textbf{to} $I$}
\State $r_{s,j} \gets \text{eval\_3d\_consistency}(\mathcal{D}(\zb_{0|t_{i+1}}^s), \Tilde{\xb}_j)$
\EndFor
\State $r_s \gets\frac{1}{I}(r_{s,1}+r_{s,2}+\cdots+r_{s,I})$
\EndFor
\State $s^* \gets \text{argmax}_s \; r_s$
\State $\zb_{t_{i+1}} \gets \zb_{t_{i+1}}^{s^*}$
\EndFor
\State $\Tilde{\xb}_{I+1} \gets \mathcal{D}(\zb_{t_T})$ \\
\Return {$\Tilde{\xb}_{I+1}$}
\end{algorithmic}
\end{algorithm}

After obtaining denoised estimates ($Z_{0|t}$ for the diffusion model, $Z_{1|t}$ for the flow model), they are decoded and compared against previously completed images using the method from~\cite{asim2025met3r} to ensure multi-view consistency. The best sample is selected for the next step. Specifically, for the initial view completion, the first two views are inpainted simultaneously by extending the given algorithm, which evaluates consistency across all possible sample pairs and selects the best pair for the next step.

\vspace{-1mm}
\section{Additional Experimental Results}
\label{sup: exp-results}

\subsection{Using Warped Videos for Fine-tuning}
\label{sup: warped-videos-user-study}
Due to the absence of an automatic metric for \textit{multi-view motion fidelity}, we conduct a human evaluation to assess the necessity of using warped videos during fine-tuning (Sec. \ref*{sec: method-training}). 
Participants were shown two randomly selected videos and asked, ''Does the generated video accurately preserve the input video's motion?" before choosing the superior video. The results are shown in Tab. \ref{appendix: tab-without-warped}.

\begin{table}[!h]
\centering
\newcommand{\first}[1]{\textbf{#1}}
\newcommand{\second}[1]{\underline{#1}}
\definecolor{Gray}{gray}{0.9}
\caption{
    \textbf{Quantitative ablation on warped videos during fine-tuning.}
    Multi-view motion fidelity is evaluated via human studies.
      }
  \begin{adjustbox}{max width=\columnwidth}
  \begin{tabular}{ccc}
    \toprule
      & w/ warped videos & w/o warped videos \\
    \hline
   Multi-view motion fidelity   & \textbf{80.44\%} & 19.56\% \\
    \bottomrule
    \end{tabular}
    \end{adjustbox}
  \label{appendix: tab-without-warped}
\end{table}


\begin{figure}[!thb]
    \centering
    \includegraphics[width=\columnwidth]
    {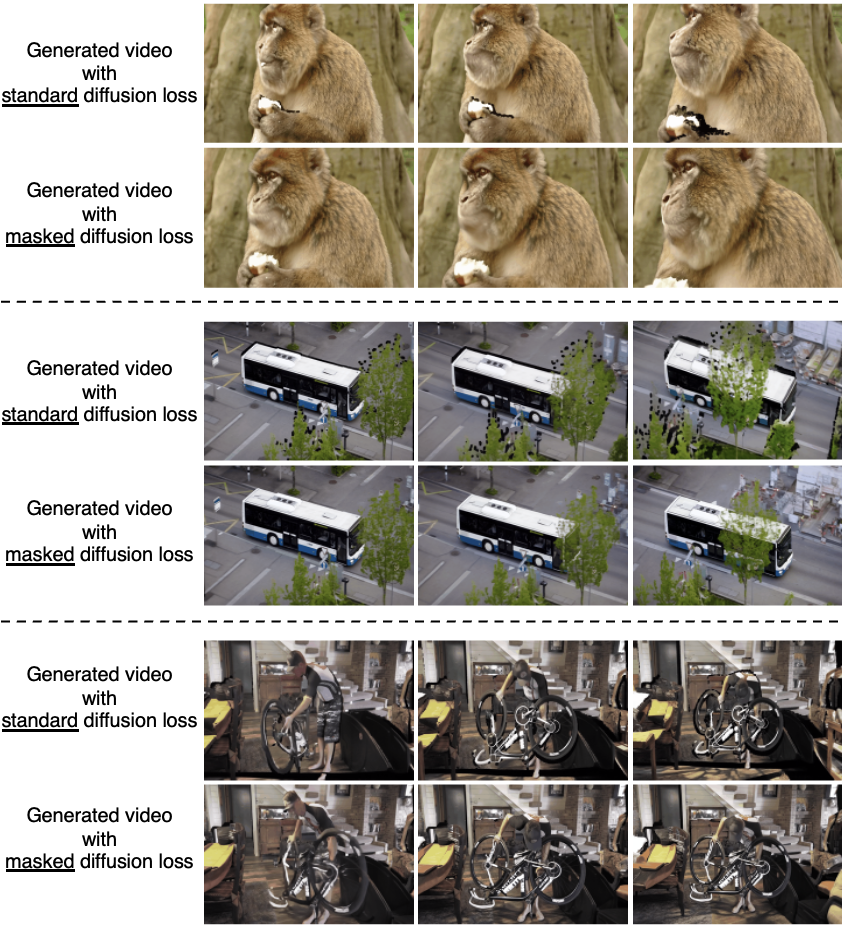}
    \caption{
    \textbf{Impact of masked diffusion loss on video quality.} 
    Masking the diffusion loss effectively prevents artifacts.
    }
    \label{appendix: masked-diffusion-loss}
    \vspace{-1.5mm}
\end{figure}

\subsection{Masking Diffusion Loss}
\label{sup: masked diffusion loss}
Building on the flexible compositionality of diffusion objectives, masked diffusion loss has been applied to diffusion-based customizations \cite{avrahami2023break, zhang2024recapture}, video interpolation \cite{chou2024generating}, and efficient image diffusion model training \cite{gao2023mdtv2, zheng2023fast}. 
In our work, we employ masked diffusion loss on a pre-trained video diffusion transformer architecture to distill the 4D motion prior of an arbitrary scene. 
As shown in Fig. \ref{appendix: masked-diffusion-loss}, this objective effectively prevents artifacts and eliminates warped (black) regions in the generated videos.


\begin{figure}[!thb]
    \centering
    \includegraphics[width=\columnwidth]
    {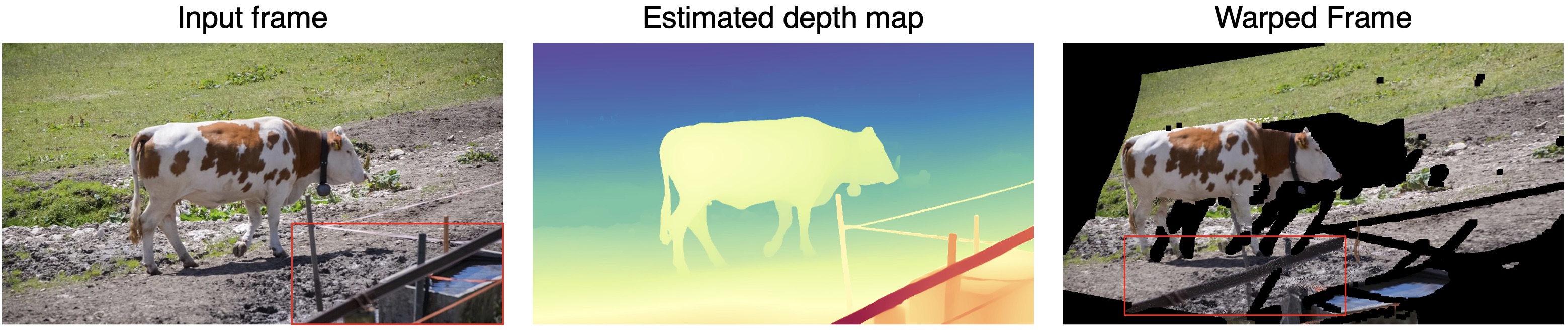}
    \caption{
    \textbf{Geometric misalignment in the warped frame.}
    In this example, the target camera view is shifted 10 degrees to the (horizontal orbit) right of the input frame.
    }
    \label{appendix: fig-limit-1}
    \vspace{-1mm}
\end{figure}

\begin{figure}[!thb]
    \centering
    \includegraphics[width=0.96\columnwidth]
    {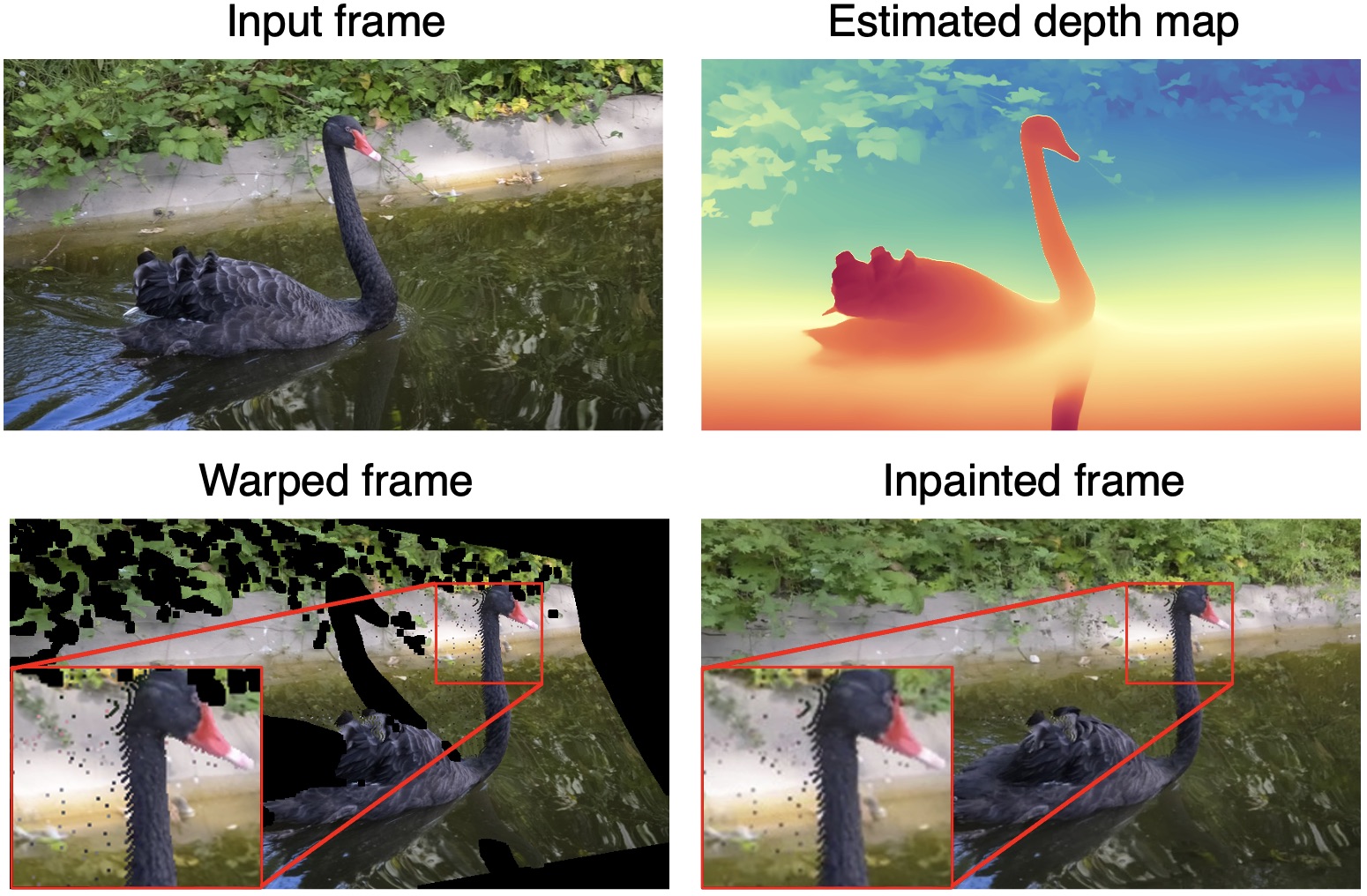}
    \caption{
    \textbf{Pixel-scale artifacts in the warped frame.}
    In this example, the target camera view is shifted 10 degrees to the (horizontal orbit) left of the input frame.
    }
    \label{appendix: fig-limit-2}
    \vspace{-1mm}
\end{figure}

\begin{figure}[!thb]
    \centering
    \includegraphics[width=0.91\columnwidth]
    {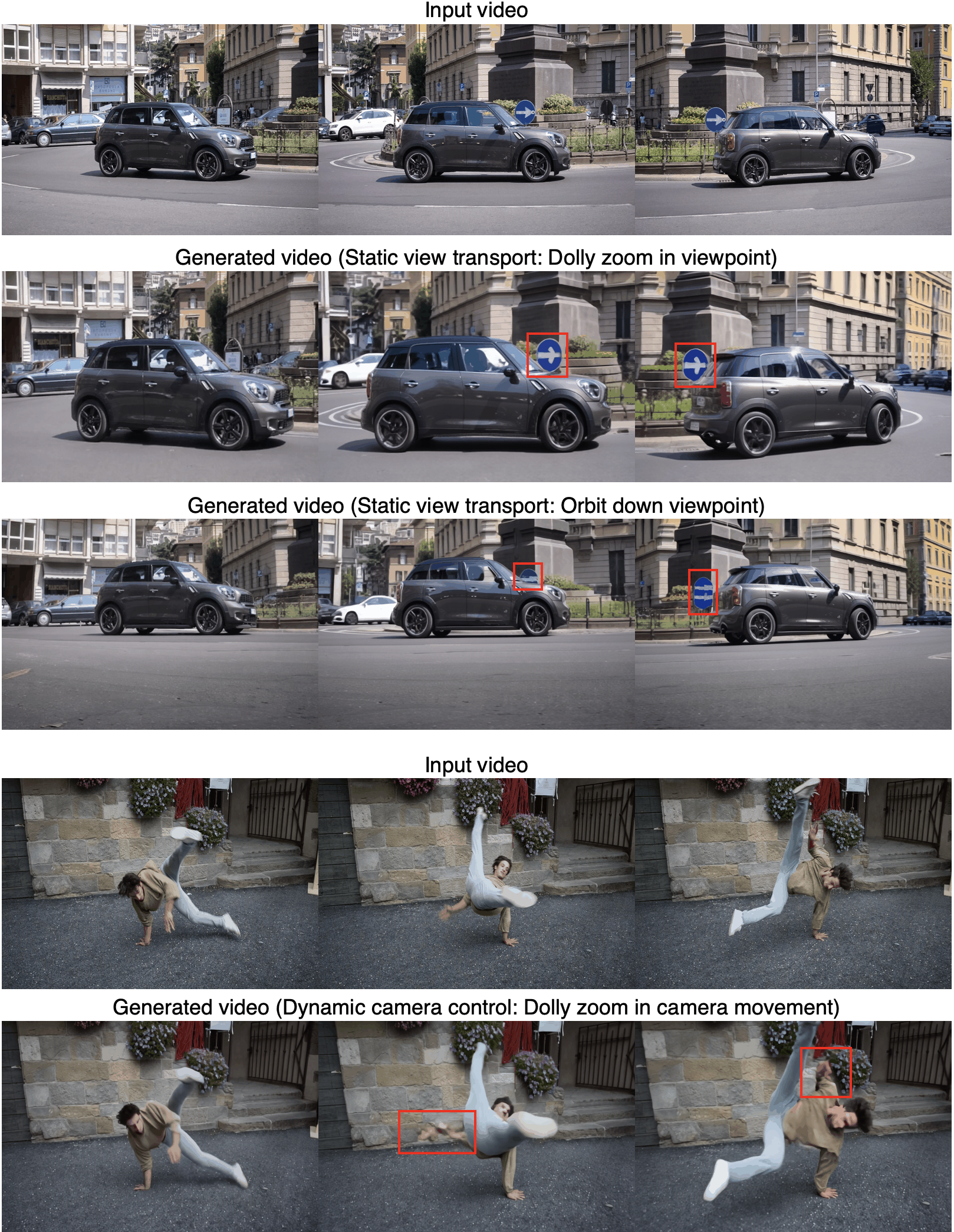}
    \caption{
    \textbf{Failure cases of Reangle-A-Video.}
    }
    \label{appendix: failure-case}
    \vspace{-2mm}
\end{figure}

\vspace{-2.5mm}
\section{Limitations and Failure Cases}
\label{sup: limitations}

Our method relies on point-based warping using estimated depth maps, making it inherently vulnerable to errors from inaccurate depth estimation and incorrect camera intrinsics. These inaccuracies can distort the warping process, leading to geometric misalignment and depth inconsistencies. For instance, Fig. \ref{appendix: fig-limit-1} illustrates an example of geometric misalignment in the warped frame.
Moreover, as shown in Fig. \ref{appendix: fig-limit-2}, warping errors can introduce pixel-level artifacts that may not be fully masked by visibility estimations, allowing these small distortions to persist in the inpainted results. Addressing these limitations would require more accurate depth estimation and better handling of occlusion constraints.
In Fig. \ref{appendix: failure-case}, we present failure cases. For instance, our method produces inaccurate reconstructions in videos with small regions (e.g., a blue sign in the background) or fast, complex motion (e.g., a man breakdancing). 
We attribute these issues partly to the fact that video fine-tuning and inference are performed in a spatially and temporally compressed latent space.

\end{document}